\documentclass[runningheads]{llncs}

\usepackage[year=2026,ID=4557]{eccv}
\usepackage[table]{xcolor}

\usepackage{eccvabbrv}
\usepackage{multirow}

\newcommand{\x}{{\bm x}}

\newcommand{\D}{\mathcal{D}}
\newcommand{\Y}{\mathcal{Y}}
\newcommand{\A}{\mathcal{A}}

\newcommand{\R}{\mathbb{R}}

\definecolor{cvprblue}{rgb}{0.21,0.49,0.74}
\definecolor{LightCyan}{rgb}{0.88,1,1}
\definecolor{Gray}{gray}{0.85}
\definecolor{myblue}{rgb}{0.88,0.98,1}

\usepackage{graphicx}
\usepackage{booktabs}
\usepackage{bm}
\usepackage{algorithmic}

\usepackage[accsupp]{axessibility}  

\usepackage{hyperref}

\usepackage{orcidlink}
\usepackage{algorithm}

\begin{document}

\title{Non-Forgetting Knowledge Allocation with Bi-level Competition for Class-Incremental Learning}

\titlerunning{NoFA-BC}

\author{
\textbf{Xiang Tan}\inst{1} \and
\textbf{Run He}\inst{1} \and
\textbf{Yawen Cui}\inst{2} \and
\textbf{Mengchen Zhao}\inst{1} \and
\textbf{Yan Wu}\inst{3} \and
\textbf{Tianyi Chen}\inst{4} \and
\textbf{Huiping Zhuang}\inst{1}\thanks{Corresponding author.} \and
\textbf{Xiaonan Luo}\inst{5} \and
\textbf{Guanbin Li}\inst{6}
}

\institute{
$^{1}$South China University of Technology \quad
$^{2}$Hong Kong Polytechnic University \quad
$^{3}$Agency for Science, Technology and Research \quad
$^{4}$Microsoft \and
$^{5}$Guilin University of Electronic Technology \quad
$^{6}$Sun Yat-sen University \\
\email{hpzhuang@scut.edu.cn}
}

\authorrunning{Tan. et al.}

\maketitle

\begin{abstract}
  Class-Incremental Learning (CIL) with pre-trained models (PTMs) aims to sequentially adapt PTMs to new categories without forgetting old knowledge. Built upon PTMs, existing adapter-based methods mainly train models via distinct task-specific adapters, and present a uniform knowledge allocation for each adapter during inference. However, this allocation mechanism ignores the nature of task discrepancy and leads to suboptimal utilization of adapters. Also, under CIL constraint, an allocator is prone to forgetting when tasks evolve. To address these issues, we propose a Non-Forgetting Allocation with Bi-Level Competition (NoFA-BC). NoFA-BC constructs a non-forgetting allocator (NFA) by transforming the allocator training into a recursive least-squares problem and achieves an allocator equivalent to that trained with all data. Based on the NFA, a Bi-Level Competition (BLC) including an intra-task level Winner-Takes-All (WTA) mechanism and inter-task Last-Ones-Fall (LOF) elimination is proposed to provide better allocation of adapter knowledge. WTA extracts the most significant logit within a task to represent the adapter's contribution and LOF suppresses the irrelevant adapters. With BLC, participation ratio of each adapter can be tailored for each input. Moreover, a Stability Enhancement (SE) process is incorporated to further improve the performance of old tasks. Extensive experiments demonstrate NoFA-BC’s superior performance across diverse datasets, consistently outperforming state-of-the-art methods.
\end{abstract}

\section{Introduction}
\label{sec:intro}

In traditional deep learning, models are often trained on a fixed dataset that is available at all times. However, {this protocol may fail in real-world scenarios, where data, e.g., new categories, appear in streaming form and are only available in certain time slots. In such circumstances, a model is required to be equipped with the capability of learning new tasks when new categories arise continually, i.e., the ability of class-incremental learning (CIL)~\cite{rebuffi2017learning, zhou2024continual}. However, CIL suffers from the issue of \textit{catastrophic forgetting}~\cite{belouadah2021comprehensive, mccloskey1989catastrophic}, where the performance of the model plunges on old tasks after learning new tasks. Consequently, to achieve good performance during the CIL, model should mitigate the forgetting issue to retain old knowledge (\textit{stability}) while maintaining the ability to attain new knowledge (\textit{plasticity})~\cite{mermillod2013stability}.} \looseness-1

\begin{figure}[t]
  \centering
  \includegraphics[width=0.6\textwidth]{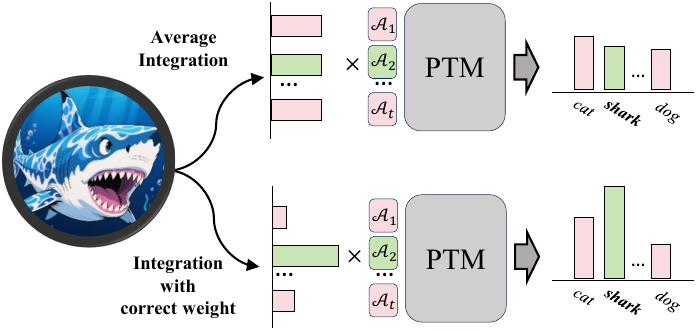}
  \caption{When aggregating multiple adapters for inference, uniform utilization of all adapters leads to insufficient application of task-relevant knowledge. Conversely, properly allocating adapters optimally leverages task-specific knowledge, thereby enhancing classification accuracy.}

  \label{fig:weight allocation}
\end{figure}

Recently, Pre-trained Models (PTMs) have demonstrated excellent performance {in downstream applications due to their generalizability~\cite{dosovitskiy2020image,thengane2022clip}. However, the forgetting issue in the CIL of PTMs still persists when PTMs are transferred to downstream tasks that continually arise~\cite{wang2022learning}. Existing PTM-based methods incorporate lightweight modules such as prompts~\cite{jia2022visual} or adapters~\cite{chenadaptformer} to adapt PTMs to new tasks, among which adapter-based methods have exhibited preferable performance~\cite{zhou2024revisiting,gao2025knowledge}. Early attempts of adapter-based methods leverage a single adapter to continually adapt PTMs during CIL~\cite{LAE2023ICCV, zhou2024revisiting}. However, they can fail to balance the stability and plasticity, and encounter a significant performance drop when substantial domain shifts exist across the tasks within CIL. To address this problem, multi-adapter strategies utilize task-specific adapters to accommodate PTMs to new knowledge within distinct subspaces and seek to aggregate task-specific knowledge \cite{zhou2024expandable,fukuda2024adapter} during the inference.}

{Although the idea of encapsulating task-specific knowledge within distinct adapters is promising, \textbf{how to effectively utilize the task-specific knowledge within multi-adapters} is a remaining problem that significantly influences the performance of adapter-based methods. One idea is to treat the adapters as independent elements and choose only one distinct adapter during inference. However, this manner ignores the potential of knowledge transfer between different tasks. On the contrary, most existing methods (e.g., ACMap \cite{fukuda2024adapter}, EASE \cite{zhou2024expandable}) seek to uniformly utilize these modules. However,}  when task distributions exhibit significant discrepancies, uniformly assigning weights to all adapters results in diminished contribution of task-specific knowledge. For example, as shown in Figure \ref{fig:weight allocation}, Adapter $\mathcal{A}_2$ (trained with data in task $2$) contains knowledge about ``sharks''. When detecting ``shark'' samples, {averagely allocating adapters weakens the contribution of $\mathcal{A}_2$, leading to knowledge interference from other adapters and misclassification. Conversely, increasing the utilization weight of $\mathcal{A}_2$ allows better extraction of image features, leading to correct classification.}  Therefore, a knowledge allocation mechanism should be established to correctly distinguish relevant adapter while suppressing irrelevant ones, thus guaranteeing the accuracy and efficiency of knowledge utilization. Moreover, given the constraint of CIL, {the allocator must be robust to forgetting during the incremental learning.}

Facing the aforementioned challenge, we propose \textbf{No}n-\textbf{F}orgetting \textbf{A}llocation with \textbf{B}i-Level \textbf{C}ompetition (NoFA-BC) in this work. 
{ NoFA-BC introduces a Non-Forgetting Allocator (NFA) with a Bi-Level Competition (BLC) mechanism to provide more effective weighting coefficient of each adapter. NFA establishes an allocator by transforming the allocation into a recursive least-squares problem and equalizes the continual allocator training to the joint learning, achieving the non-forgetting for allocating. Based on NFA, BLC leverages the logits generated by the NFA to perform both intra-task level and inter-task level competition. A {Winner-Takes-All} (WTA) mechanism is proposed for the intra-task level to extract the most representative information related to the adapters to determine their contribution and a Last-Ones-Fall (LOF) mechanism is presented to eliminate the negative effect of irrelevant adapters for inter-task level. With BLC, participation ratio of each adapter can be dynamically tailored for each input sample under the principle that distinguishing relevant adapter while suppressing irrelevant ones. Moreover, we observe the good potential of stability within the NFA and propose a Stability Enhancement (SE) process to further improve the stability of our method. By incorporating all these modules,} NoFA-BC achieves state-of-the-art performance across multiple benchmark datasets. Specifically, NoFA-BC demonstrates superior performance on Task 40 of ImageNet-R (long-sequence scenario) and Task 10 of ImageNet-A (high-complexity scenario), surpassing the second-best method by 4.09$\%$ and 7.87$\%$ in accuracy.

Our key contributions are summarized as follows:
{
\begin{itemize}
    \item We propose the NoFA-BC, a novel adapter-based method that better leverages the knowledge encoded in multi-adapters for CIL with PTMs.
    \item To obtain an allocator robust to forgetting, NFA is established via a recursive least-squares solution, enabling a non-forgetting allocator.
    \item To effectively manage the contribution of each adapter, based on the NFA, BLC introduces a WTA and an LOF mechanism for knowledge allocation, greatly enhancing knowledge utilization.
    \item An SE process is designed to leverage the stability within the NFA, further improving the model stability and overall performance. 
    \item Extensive experiments across multiple benchmark datasets and settings are conducted, demonstrating that NoFA-BC achieves satisfactory knowledge allocations and outperforms state-of-the-art (SOTA) methods.
\end{itemize}
}

\section{Related Work}
\label{sec:related}

\noindent\textbf{Class-Incremental Learning (CIL).} Traditional class-incremental learning { aims to sequentially learns different categories based on a backbone without pre-training, and it can be broadly categorized into three paradigms.} Replay-based methods~\cite{rebuffi2017icarl, liu2020mnemonics,zhao2021memory} effectively mitigate catastrophic forgetting {by storing and replaying a subset of historical data during CIL. However, this technique violates data privacy to some extent and introduce further storage overhead}. Knowledge distillation-based approaches \cite{LwF2017TPAMI,kirkpatrick2017overcoming} {incorporate additional regularization terms into the loss function to restrict parameter change and such a constraint can cause poor plasticity.} Model expansion-based~\cite{chen2023dynamic,hu2023dense,wang2022foster,yan2021dynamically} methods employ a dedicated backbone network for each task while freezing and saving parameters from previous backbones during training to keep previous knowledge, which {leads to large storage overhead. }

\noindent\textbf{PTM-Based CIL.} {PTM-based CIL have emerged as a new branch in class-incremental learning. With PTMs, the CIL is conducted using parameter-efficient tuning (PET) methods including prompts and adapters. For example,} L2P~\cite{wang2022learning} introduces a key-query matching mechanism to select an optimal prompt. DualPrompt~\cite{wang2022dualprompt} further differentiates between G-Prompt (handling shared knowledge) and E-Prompt (addressing task-specific knowledge). CODA-Prompt~\cite{smith2023coda} employs multiple trainable prompts with a dynamic weighting mechanism to determine their relative contributions during inference.  APER~\cite{zhou2024revisiting} first revealed the remarkable effectiveness of prototype-based classification with only fine-tuning with PET on the first task. EASE~\cite{zhou2024expandable} constructs a distinct subspace for each task via adapters and estimates prototype representations of historical classes based on similarity within the current subspace. Additionally, ACMap~\cite{fukuda2024adapter} fuses current and historical adapters in parameter space to share knowledge. However, a more effective way for knowledge allocation remains underexplored in adapter-based CIL.

\noindent\textbf{CIL with Least-Squares Solutions.} The least-squares method has recently exhibited significant potential in CIL~\cite{FOAL2024NeurIPS,GACL2024NeurIPS,he2024real}. Inspired by pseudo-inverse learning~\cite{guo2004pseudoinverse}, ACIL~\cite{zhuang2022acil} first proves that, when equipped with a frozen feature-extraction backbone, recursive least squares can achieve performance equivalent to joint training, showing competitive performance in CIL. DS-AL~\cite{zhuang2024ds} introduces a compensation module to address underfitting issue during training. AIR~\cite{fang2024air} incorporates a class-wise weighting parameter into recursive least squares computations to solve the problem of imbalanced training data. DPCR~\cite{he2025semantic} addresses both semantic shift and decision bias using least-squares. Additionally, least squares methods show great performance in multi-modal CIL~\cite{xu2024advancing, yue2024mmal}, few-shot CIL~\cite{GKEAL2023CVPR}, multi-label CIL~\cite{zhang2025l3a}, and other CIL-related applications~\cite{HSIC2025JFI, AEF-OCLTVT2025, CrossACL2025GRSL}. In this paper, we leverage the non-forgetting property of recursive least-squares to construct an allocator for adapters.

\section{Preliminaries}
\label{sec:preliminaries}

\subsection{Class-Incremental Learning}

In Class-incremental learning scenarios, models are required to sequentially acquire knowledge across $T$ distinct tasks, formally represented as a task sequence $\{\D_1,\D_2,\cdots,\D_T\}$. The $t$-th task $\D_t = {\{(\x_{t,i}, y_{t,i})\}}_{i=1}^{n_t}$, comprises $n_t$ data pairs. Each pair consists of input data $\x_{t,i} \in \mathcal{X}_t$ and a label $y_{i}^{t} \in \Y_{t}$, where $\mathcal{X}_t$ is the sample set and  $\Y_{t}$ is the label set in task $t$. The model at task $t$ can thus be formulated as a parametric mapping function $f_{\theta} : \mathcal{X}_{1:t} \rightarrow \Y_{1:t}$ for all seen classes, where $\theta$ denotes the trainable parameters.

Under CIL protocols, label sets across each different tasks $t$ and $t^{\prime}$ are  strictly disjoint, i.e., $\Y_t \cap \Y_{t^{\prime}}=\varnothing$. Furthermore, considering the data privacy, we follow {exemplar-free setting \cite{masana2022class}, in which data of previous tasks} $\D_{1:t-1}$ remains inaccessible during the training phase $t$. Consequently, only the current task data $\D_{t}$ participates in parameter updates at task $t$. \looseness-1

\begin{figure*}[t]
	\begin{center}
		\includegraphics[width=\textwidth]{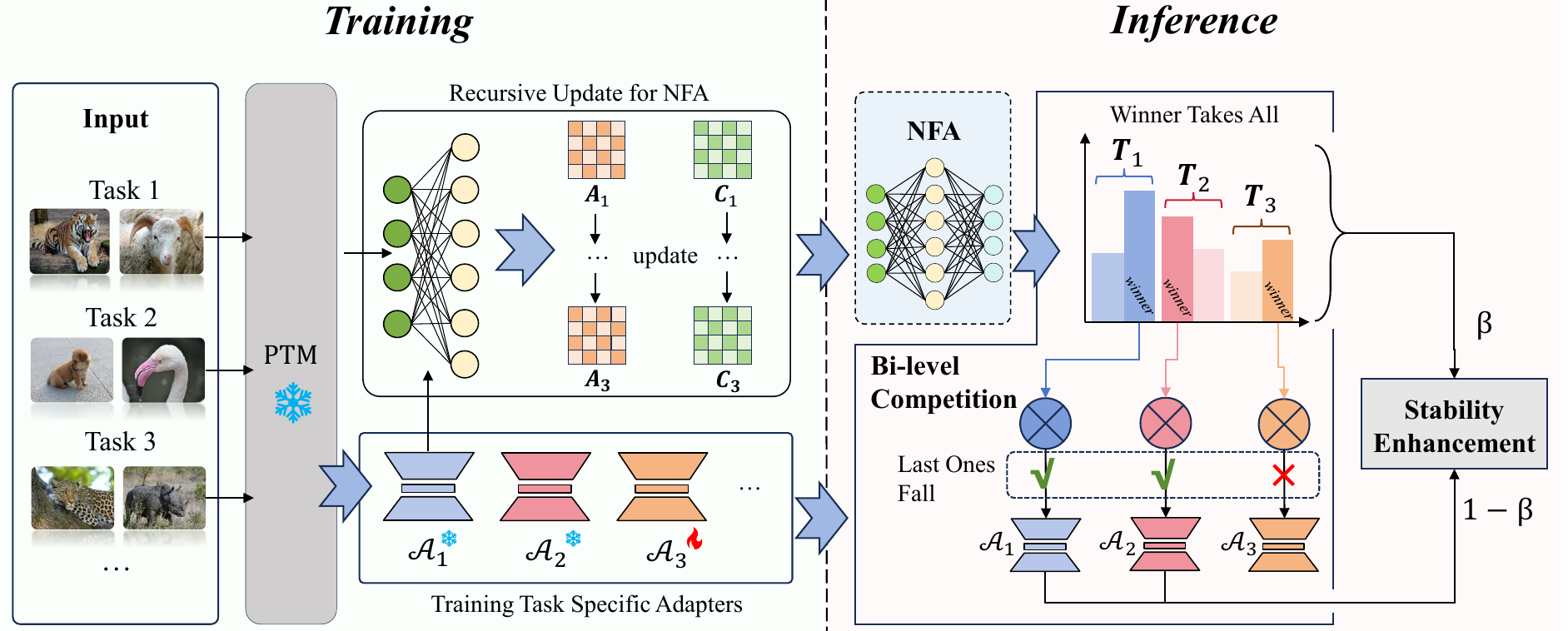}
	\end{center}
	\caption{\small The Overview of NoFA-BC. {\bf Left}: Training Phase, assuming current task is $3$. Task-specific adapters are trained per incremental stage while the autocorrelation and cross correlation matrix is continuously updated with the help of adapter $\mathcal{A}_1$.
	{\bf Right}: The NoFA-BC dynamically regulates the participation of each adapters with the help of Bi-level Competition mechanism. SE is utilized to enhance the stability of final output.} 
    \label{figure:teaser}
\end{figure*} 

\subsection{Adapter-based CIL}
{Existing adapter-based CIL mainly utilizes a a pre-trained Visual Transformer (ViT) as backbone and incorporates lightweight adapters to adapt the backbone to new data. During the CIL, the backbone is frozen and only the adapters are updated. Suppose there are $L$ blocks in the ViT, each with a self-attention module and an MLP layer, the adapter can be included in the MLP in parallel. An adapter is a bottleneck structure consisting of a down-projection layer $W_{down} \in \R^{d\times r}$, a non-linear activation function $\sigma$ (ReLU in general), and an up-projection layer $W_{up} \in \R^{d\times r}$, where $r$ is the bottleneck dimension}. Denote the input token $x_{in} \in \R^{d}$ and output token $x_{out}\in \R^{d}$ are the input and output of the MLP layer, {the forward process of the MLP layer with the adapter can be formalized as:}

\begin{align} 
\label{eq:adapter-define}
	\x_{\text{out}} = \text{MLP}(\x_{\text{in}})+\sigma(\x_{\text{in}} \boldsymbol{W}_{\text{down}})\boldsymbol{W}_{\text{up}} .
\end{align}

In practice, {an adapter can be incorporated into each layer of a ViT}. In this paper, we denote all adapters inserted into the pre-trained model as one adapter {set $\A$}. In each task, we train a task-specific adapter set to learn the knowledge of coming data. {For task $t$, the corresponding adapter is denoted as $\A_t$, and the feature (i.e., the cls token of last block) processed with $\A_t$ and the original ViT is denoted as $\phi(\x, \A_t)$. }

\section{Proposed Method}
\label{sec:NoFA-BC}

{In this section, we present the details of the proposed NoFA-BC. Similar to existing multi-adapter methods, NoFA-BC trains independent adapters for different tasks but seeks to effectively utilize the task-specific knowledge. To achieve this, NoFA-BC first conducts an Integration of Adapters (IA) process during inference to obtain knowledge transfer via prototypes across tasks, and then leverages a Non-Forgetting Allocator (NFA) with Bi-Level Competition (BLC) mechanism for allocating these knowledge. Finally, a Stability Enhancement (SE) process is incorporated to further improve the stability of model.}


\subsection{Integration of Adapters} \label{sec:merge}
To encapsulate task-specific knowledge into subspaces formed by the adapters, we follows previous works to train an adapter for each task. To achieve better knowledge transfer across each task, here we combine all adapters to use knowledge from all stages via task-related prototypes. In task $t$, NoFA-BC averages embeddings in feature space, getting averaged features from each adapter’s inference output:
\begin{align} \label{eq:Integration of Adapters0}
    \Phi_t(\boldsymbol{x}) = \frac{\sum_{i=1}^{t}\phi(\boldsymbol{x},\A_i)}{t}.
\end{align}

After the training of each task, we use the training dataset of the current task to construct a prototypical classifier. Specifically, we employ the integration strategy in Eq.\ref{eq:Integration of Adapters0} to extract the prototypes of all categories and calculate their average values:
\begin{align} \label{eq:prototypical classifier}
    \boldsymbol{p}_{c}=\frac1{N_c}\sum\nolimits_{j=1}^{|\mathcal{D}^t|}\mathbb{I}(y_j=c)\Phi_t(\boldsymbol{x}_j) ,
\end{align}
where $N_c$ is the sample number of class $c$, $\mathbb{I}(\cdot)$ is the indicator function. These prototypes are used as the weight of the classifier, which could be represented as $\boldsymbol{W}^P=[\boldsymbol{p}_{1},...,\boldsymbol{p}_{c}]\in \R^{C \times d}$, where $C$ represents the total number of categories in dataset. During inference, the classifier calculates the cosine similarity between $W$ and $\Phi_t(\boldsymbol{X})$ for classification: 
\begin{align} \label{eq:cosine similarity}
    f(\boldsymbol{x})=(\frac {\boldsymbol{W}^P}{\|\boldsymbol{W}^P\|_2})^\top(\frac{\Phi_t(\boldsymbol{x})}{\|\Phi_t(\boldsymbol{x})\|_2}).
\end{align}



\subsection{Non-Forgetting Allocation with Competition} \label{sec:refine}
{The operation of IA is a direct way for knowledge transfer, but the critical issue of knowledge interference if uniformly utilizing adapters still remains. A simple technique is to train a learnable allocator to adjust the contributions of each adapter. However, directly using gradient-based optimization without access to previous data will lead to catastrophic forgetting of the allocator \cite{FOAL2024NeurIPS, smith2023coda}. Also, utilizing all adapters via task-level information may not be optimal due to the diversity of categories within a task. Consequently, we propose a Non-Forgetting Allocator with a Bi-Level Competition mechanism for allocating adapters.}
\looseness-1

{
\textbf{Non-Forgetting Allocator.} To develop a non-forgetting allocator, we transform the allocation process into a recursive least-squares problem with features extracted from a frozen backbone. This process ensures that the allocator trained in a sequential manner achieves identical results to that trained by jointly collected data, i.e., the Non-Forgetting Allocator. 
Inspired by APER~\cite{zhou2024revisiting}, in order to bridge the domain gap between the PTM and downstream datasets, we incorporate the adapter trained in the first task to the backbone for all subsequent tasks. Subsequently, a training-free random buffer $f_{B}(\cdot):\R^d\rightarrow \R^{d_{B}}(B>d)$ is applied to enhance the feature representation $\phi(\boldsymbol{x}, \A_1)$. This process non-linearly maps the features to a higher-dimensional space and enhances separability across features \cite{Cover1965TEC, zhuang2022acil} via a fixed random feature expansion matrix $\boldsymbol{B}$. For the stacked features in task $t$, it can be represented as
\begin{equation} \label{eq:buffer layer}
    \boldsymbol{X}_t =  f_{\text{stack}}(f_{B}(\phi(\boldsymbol{x}_t,\A_1))) = f_{\text{stack}}(\sigma_{B}(\phi(\boldsymbol{x}_t,\A_1)\boldsymbol{B})),
\end{equation} 
where $\sigma_{B}$ is a non-linear activation function and $f_{\text{stack}}$ denotes stacking all the features of inputs in task $t$. }


{The primary objective for allocation is to utilize the most significant task-specific information. However, within the same task, the distribution differences between categories can be significant. Similarly, categories in different tasks may be relatively close. Such a distribution can result in poor discriminative performance across tasks. Therefore, we establish the allocator based on} class-discriminative information by minimizing the following objective function:
\begin{equation} \label{eq:opt1}
    \underset{\boldsymbol{W}_t^A}{\operatorname*{\operatorname*{argmin}}}\|\boldsymbol{X}_{1:t}\boldsymbol{W}_t^A-\boldsymbol{Y}_{1:t}\|_{\mathrm{F}}^2+\gamma\|\boldsymbol{W}_t^A\|_{\mathrm{F}}^2,
\end{equation}
where $\|\cdot\|_{\mathrm{F}}^2$ represents the Frobenius-norm and $\gamma$ is regularization {factor}. $X_{1:t}$ and ${Y}_{1:t}$ indicate the {stacked feature expansion matrix from task $1$ to $t$ obtained by Eq. \eqref{eq:buffer layer} and corresponding one-hot labels, formulated as follow}
\begin{equation}\boldsymbol{X}_{1:t}=
\begin{bmatrix}
\boldsymbol{X}_{1:t-1} \\
\boldsymbol{X}_t
\end{bmatrix},\quad\boldsymbol{{Y}}_{1:t}=
\begin{bmatrix}
\boldsymbol{{Y}}_{1:t-1}\\
  \boldsymbol{Y}_t
\end{bmatrix}.\end{equation}

{The optimal solution of Eq. \eqref{eq:opt1} is}
\begin{equation}
\label{eq:W}
\boldsymbol{{W}}_t^A=\left(\sum^t_{i=1}\boldsymbol{X}_{i}^\top\boldsymbol{X}_{i}+\gamma\boldsymbol{I}\right)^{-1}\left(\sum^t_{i=1}\boldsymbol{X}_{i}^\top\boldsymbol{Y}_{i}\right).\end{equation}

Through Eq. \ref{eq:W}, we obtain the weight of the allocator $\boldsymbol{{W}}_t^A$ for phase $t$. However, due to the exemplar-free constraints, we cannot access historical data $\boldsymbol{X}_{1:t-1}$ from all previous tasks. {To eliminate the reliance on previous data, we transform this process into a recursive manner. Suppose the autocorrelation matrix ${\boldsymbol{A}}_t$ and the cross correlation matrix ${\boldsymbol{C}}_t$ can be obtained by}
\begin{equation}
    \label{AC}{\boldsymbol{A}}_t=\sum^t_{i=1}\boldsymbol{X}_{i}^\top\boldsymbol{X}_{i},\quad
    {\boldsymbol{C}}_t=\sum^t_{i=1}\boldsymbol{X}_{i}^\top\boldsymbol{Y}_{i}, 
\end{equation}
{the update of $\boldsymbol{{W}}_t^A$ can be transformed into the recursive update of ${\boldsymbol{A}}_t$ and ${\boldsymbol{C}}_t$ from the Eq. \ref{AC}. ${\boldsymbol{A}}_{t+1}$ and ${\boldsymbol{C}}_{t+1}$ can be calculated by}
\begin{equation}\label{updateAC}
\begin{split}
    &{\boldsymbol{A}}_{t+1}={\boldsymbol{A}}_{t}+\boldsymbol{X}_{t+1}^\top\boldsymbol{X}_{t+1},\\
    &{\boldsymbol{C}}_{t+1}={\boldsymbol{C}}_{t}+\boldsymbol{X}_{t+1}^\top\boldsymbol{Y}_{t+1}. 
\end{split}
\end{equation}

Therefore, the update of $\boldsymbol{{W}}_t^A$ can be written as
\begin{equation}
\label{eq:updateW}
\begin{split}
    &\boldsymbol{{W}}_{t+1}^A =\left(\sum^{t+1}_{i=1}\boldsymbol{X}_{i}^\top\boldsymbol{X}_{i}+\gamma\boldsymbol{I}\right)^{-1}\left(\sum^{t+1}_{i=1}\boldsymbol{X}_{i}^\top\boldsymbol{Y}_{i}\right)\\
    &=\left({\boldsymbol{A}}_{t+1}+\gamma\boldsymbol{I}\right)^{-1}{\boldsymbol{C}}_{t+1}\\
    &=\left({\boldsymbol{A}}_{t}+\boldsymbol{X}_{t+1}^\top\boldsymbol{X}_{t+1}+\gamma\boldsymbol{I}\right)^{-1}\left({\boldsymbol{C}}_{t}+\boldsymbol{X}_{t+1}^\top\boldsymbol{Y}_{t+1}\right).
\end{split}
\end{equation}

As shown in Eq. \ref{eq:updateW}, {the solution to the joint learning problem in task $t+1$ can be transformed into a recursive update with only the $\boldsymbol{A}_t$, $\boldsymbol{C}_t$ and the data in current task. This process enables a non-forgetting allocator which can have a identical results to that trained with all the previous data. During the CIL, we can only store the autocorrelation matrix $\boldsymbol{A}_t$ and the cross correlation matrix $\boldsymbol{C}_t$ to obtain a robust allocator after the training phase of each task $t$.} 


{
\textbf{Bi-Level Competition for Allocation.} To obtain an effective utilization of each adapter, the output of NFA should be converted into weights that assigns contribution of each adapter. If the input is more related to a distinct adapter, the contribution of the knowledge from that adapter should be raised and other non-related adapters should be suppressed. To achieve this, we propose a bi-Level competition mechanism including the \textit{Winner Takes All} (WTA) within the intra-task level and the \textit{Last Ones Fall} (LOF) for the inter-task level. 

For the intra-task level, we seek to obtain the most significant task-relevant information to represent the contribution of the adapters. Since the categories within a task can vary drastically, averagely considering the contribution of each category may mislead the allocation. In such a situation, a class-wise competition should be conducted to determine the contribution level. Here, a WTA mechanism, which selects the highest logit of classes in task $i$ in the output as the weight coefficient is established as follow }
\begin{equation}\label{eq:alpha}
    \alpha_i = \mathop{\max}\limits_{k \in \mathcal{Y}_i} \ \text{Softmax}(f_{B}(\phi(\boldsymbol{x},\A_i))\boldsymbol{W}_t^A)[k].
\end{equation}

In this way, the most representative weights of the task are extracted and {a coefficient set for allocating each adapter ${\{\alpha_1,\alpha_2,...,\alpha_t\}}$ can be obtained. However, it is still possible that knowledge learned by an adapter is not related to the input despite the weighting coefficient is low. If such an adapter is integrated, it may inject harmful knowledge for the classification. To address this, an LOF mechanism for the inter-task level is designed to eliminate the less relevant adapters. After obtaining the coefficient set, we rank the weight coefficients of each task and \textbf{set the lowest $o$-percent adapters to $0$}, only taking the significant ones.}

Then, we use the obtained weight coefficients ${\{\alpha_1,\alpha_2,...,\alpha_t\}}$ to allocate the participation of each adapter in the integration of adapters. Hence, Eq. \ref{eq:Integration of Adapters0} should be rewritten by assigning weights to the adapters:
\begin{align} \label{eq:Integration of Adapters}
    \Phi_t(\boldsymbol{x}) = \frac{\sum_{i=1}^{t}\alpha_i\cdot\phi(\boldsymbol{x},\A_i)}{t}.
\end{align}

\begin{table}[t]
	\caption{\small The average accuracy $\bar{A}$ and the last-task accuracy $A_T$. CIFAR refers to CIFAR-100 and IN-R/A stands for ImageNet-R/A. The best performance is shown in \textbf{bold} and the second best performance is shown in \underline{underline}. 
	}\label{tab:benchmark}
	\centering
    \setlength{\tabcolsep}{1mm}
    \footnotesize
	\resizebox{\textwidth}{!}{
		\begin{tabular}{p{2.0cm} cccccccccc} 
			\toprule
			\multicolumn{1}{l}{\multirow{2}{*}{Method}} & 
			\multicolumn{2}{c}{CIFAR B0 Inc5} & \multicolumn{2}{c}{IN-R B0 Inc5} 
			& \multicolumn{2}{c}{IN-R B0 Inc20}
			& \multicolumn{2}{c}{IN-A B0 Inc20}
			& \multicolumn{2}{c}{VTAB B0 Inc10} \\
			& {$\bar{{A}}$} & ${{A}_T}$  
			& {$\bar{{A}}$} & ${{A}_T}$  
			& {$\bar{{A}}$} & ${{A}_T}$ 
			& {$\bar{{A}}$} & ${{A}_T}$ 
			& {$\bar{{A}}$} & ${{A}_T}$ 
			\\
			\midrule
			Finetune	
            & 38.90{$_{\pm{0.09}}$} 
            & 20.17{$_{\pm{0.16}}$}
            &21.61{$_{\pm{1.68}}$}
            & 10.79{$_{\pm{2.48}}$}
            &32.31{$_{\pm{0.98}}$}
            &22.78{$_{\pm{1.26}}$}
            &24.28{$_{\pm{1.34}}$}
            & 14.51{$_{\pm{1.66}}$}
            & 34.95{$_{\pm{0.71}}$}
            & 21.25{$_{\pm{0.82}}$}  \\
		      Adapter-FT \cite{zhou2024revisiting}
            & 60.51{$_{\pm{0.73}}$}
            &49.32{$_{\pm{0.62}}$}
            & 47.59{$_{\pm{0.85}}$}
            &40.28{$_{\pm{1.00}}$}
            &58.17{$_{\pm{0.50}}$}
            &52.39{$_{\pm{0.37}}$}
            &45.41{$_{\pm{0.77}}$}
            &41.10{$_{\pm{0.89}}$}
            &48.91{$_{\pm{0.23}}$}
            & 45.12{$_{\pm{0.20}}$} \\
			LwF \cite{LwF2017TPAMI}
            & 46.29{$_{\pm{0.28}}$}
            & 41.07{$_{\pm{0.33}}$}
            &39.93{$_{\pm{0.14}}$}
            & 26.47{$_{\pm{0.22}}$}
            &45.72{$_{\pm{0.09}}$}
            &34.17{$_{\pm{0.12}}$}
            &37.75{$_{\pm{0.81}}$}
            & 26.84{$_{\pm{0.90}}$}
            & 40.48{$_{\pm{0.07}}$}
            & 27.54{$_{\pm{0.10}}$}\\
			L2P \cite{wang2022learning}
            & 85.94{$_{\pm{0.48}}$} 
            & 79.93{$_{\pm{0.33}}$} 
            &52.17{$_{\pm{0.22}}$} 
            & 49.20{$_{\pm{0.15}}$} 
            & 75.46{$_{\pm{0.47}}$} 
            & 69.77{$_{\pm{0.46}}$} 
            &  49.39{$_{\pm{0.99}}$} 
            & 41.71{$_{\pm{1.06}}$} 
            &  77.11{$_{\pm{0.34}}$} 
            & 77.10{$_{\pm{0.29}}$} \\
			DualPrompt \cite{wang2022dualprompt}
            &87.87{$_{\pm{0.30}}$}
            & 81.15{$_{\pm{0.53}}$}
            & 63.31{$_{\pm{0.24}}$}
            & 55.22{$_{\pm{0.45}}$}
            &73.10{$_{\pm{0.19}}$}
            &67.18{$_{\pm{0.55}}$} 
            & 53.71{$_{\pm{0.64}}$}
            & 41.67{$_{\pm{0.86}}$} 
            & 83.36{$_{\pm{0.17}}$}
            & 81.23{$_{\pm{0.21}}$}\\
			CODA \cite{smith2023coda}
            & 89.11{$_{\pm{0.99}}$}
            & 81.96{$_{\pm{1.43}}$}
            & 64.42{$_{\pm{0.59}}$}
            & 55.08{$_{\pm{0.68}}$}
            & 77.97{$_{\pm{0.58}}$}
            &72.27{$_{\pm{0.36}}$}
            & 53.54{$_{\pm{0.49}}$}
            & 42.73 {$_{\pm{0.83}}$}
            & 83.90{$_{\pm{0.67}}$}
            &83.02{$_{\pm{0.37}}$}\\
			SimpleCIL \cite{zhou2024revisiting} 
            &  87.57{$_{\pm{0.03}}$}
            & 81.26{$_{\pm{0.05}}$}
            & 62.58{$_{\pm{0.02}}$}
            & 54.55{$_{\pm{0.08}}$}
            &61.26{$_{\pm{0.05}}$}
            &54.55{$_{\pm{0.07}}$}
            & 59.77{$_{\pm{0.08}}$}
            & 48.91{$_{\pm{0.13}}$}
            & 85.99{$_{\pm{0.02}}$}
            & 84.38{$_{\pm{0.04}}$}\\
            ACIL \cite{zhuang2022acil}
            &89.88{$_{\pm{0.04}}$} &83.56{$_{\pm{0.05}}$} 
            &75.16{$_{\pm{0.18}}$} 
            &67.33{$_{\pm{0.21}}$} 
            &75.17{$_{\pm{0.05}}$}  &68.40{$_{\pm{0.22}}$} 
            &63.30{$_{\pm{0.42}}$} &53.26{$_{\pm{0.66}}$} 
            & 90.93{$_{\pm{0.12}}$}
            & 89.84{$_{\pm{0.09}}$}\\
            DS-AL \cite{zhuang2024ds}
            & 90.24{$_{\pm{0.35}}$}
            &84.12{$_{\pm{0.41}}$}
            &74.59{$_{\pm{0.64}}$}
            &67.57{$_{\pm{0.81}}$}
            &73.88{$_{\pm{0.56}}$}
            &68.13{$_{\pm{0.33}}$}
            &56.78{$_{\pm{0.48}}$}
            &53.13{$_{\pm{0.69}}$}
            &  89.48{$_{\pm{0.73}}$}
            & 88.17{$_{\pm{0.87}}$}\\
			APER \cite{zhou2024revisiting}
            &   90.65{$_{\pm{0.06}}$}
            &  85.15{$_{\pm{0.15}}$}
            &75.82{$_{\pm{0.11}}$}
            &67.95{$_{\pm{0.07}}$}
            &75.82{$_{\pm{0.03}}$}
            &67.95{$_{\pm{0.08}}$}
            & 60.47{$_{\pm{0.72}}$}
            &49.37{$_{\pm{0.66}}$}
            &  85.95{$_{\pm{0.05}}$}
            & 84.35{$_{\pm{0.04}}$}\\
            EASE \cite{zhou2024expandable}
            & 91.51{$_{\pm{0.33}}$}
            & 85.80{$_{\pm{0.52}}$}
            & \underline{78.31}{$_{\pm{0.37}}$} 
            &\underline{70.58}{$_{\pm{0.28}}$}
            & \underline{81.74}{$_{\pm{0.32}}$}
            & \underline{76.17}{$_{\pm{0.19}}$}
            & \underline{65.34}{$_{\pm{1.08}}$}
            & 55.04{$_{\pm{1.25}}$}
            & \underline{93.61}{$_{\pm{0.21}}$}
            & \underline{93.55}{$_{\pm{0.13}}$} \\
            ACMap \cite{fukuda2024adapter}
            & \underline{92.04}{$_{\pm{0.29}}$}
            & \underline{87.81}{$_{\pm{0.19}}$}
            & 77.31{$_{\pm{0.43}}$}
            &70.49{$_{\pm{0.98}}$}
            & 79.94{$_{\pm{0.59}}$}
            & 74.28{$_{\pm{0.75}}$}
            & 65.19{$_{\pm{0.79}}$}
            & \underline{56.19}{$_{\pm{0.98}}$}
            & 91.21{$_{\pm{0.26}}$}
            & 87.56{$_{\pm{0.38}}$} \\
            \midrule
                \rowcolor{gray!10}
			\textbf{NoFA-BC}
            & \raggedright \bf94.04{$_{\pm{0.33}}$}  
            & \bf89.21{$_{\pm{0.45}}$} 
            & \bf82.44{$_{\pm{0.27}}$} 
            & \bf74.67{$_{\pm{0.45}}$}  
            & \bf83.61{$_{\pm{0.37}}$}  
            & \bf77.75{$_{\pm{0.29}}$}   
            & \bf72.51{$_{\pm{0.87}}$}  
            & \bf64.06{$_{\pm{0.48}}$}   
            & \bf93.98{$_{\pm{0.25}}$}  
            & \bf{93.91}{$_{\pm{0.66}}$}   \\
			\bottomrule
		\end{tabular}
	} 
\end{table}

This approach enables the model to dynamically adjust the participation ratios of adapters. The task-specific adapter can participate more dominantly during inference for the corresponding task, correcting the final output toward proper task knowledge. In addition, other adapters contribute at varying ratios and less relevant adapters are excluded, ensuring further satisfactory allocation.

\subsection{Stability Enhancement} \label{sec:Complementation}
Maintaining stability is a prevalent challenge in CIL scenarios. Conventionally, updates or expansions of model/modules (e.g., IA in our method) embody plasticity through continual knowledge acquisition, while stability typically corresponds to freezing the backbone or module. Crucially, our NFA inherently prioritizes stability due to its frozen backbone with good generalization and non-forgetting properties. {Here we present a Stability Enhancement (SE) process to leverage stability in NFA via fusing NFA's class-relevant output with IA's result,}
\begin{equation} \label{eq:Complementary}
    \hat{y} = \beta\cdot {f_{B}(\phi(\boldsymbol{x},\A_1))}\boldsymbol{W}^A +(1-\beta) \cdot f(\boldsymbol{x}),
\end{equation}
where $\beta$ is a hyperparameter to balance the stability and plasticity and $\boldsymbol{W}^A$ is the classification weight of NFA. The pseudo-code of proposed NoFA-BC is summarized in Algorithm 1 in Supplementary Materials A.

\looseness-1

\begin{figure}[t]
	\centering
    \includegraphics[width=\textwidth]{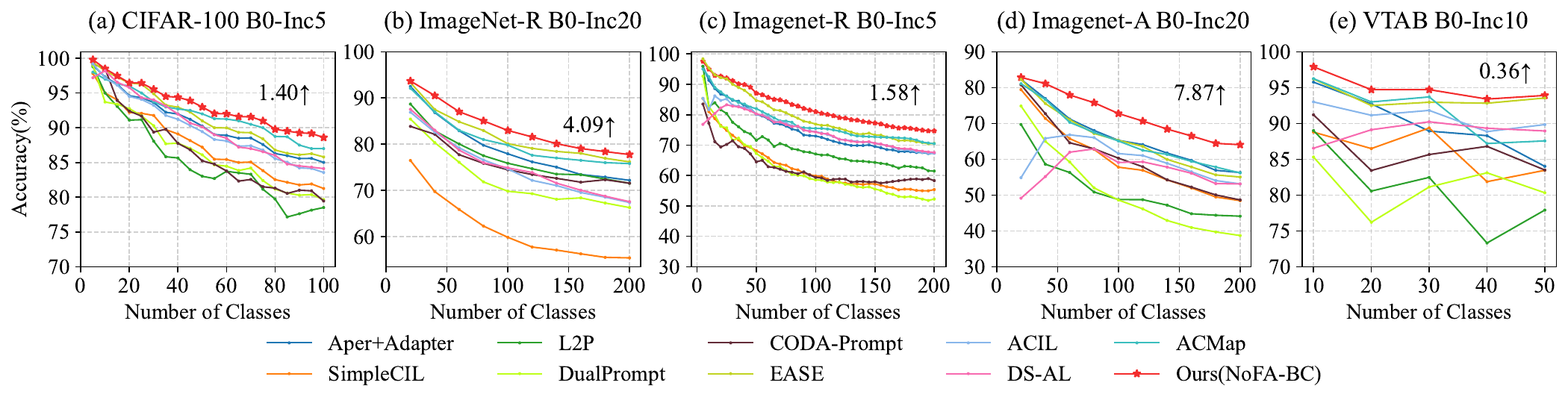}

	\caption{\small Accuracy curve of compared methods during CIL. We annotate the relative improvement of NoFA-BC above the runner-up method with numerical numbers at the last incremental stage.}
	\label{fig:benchmark}

\end{figure}

\section{Experiments}
\label{sec:exp}

{\subsection{Experiment Setting}
\noindent\textbf{Dataset and CIL Protocol.}} We follow the setting of \cite{fukuda2024adapter} to evaluate our method on four benchmark datasets: CIFAR100~\cite{krizhevsky2009learning}, ImageNet-R~\cite{hendrycks2021many}, ImageNet-A~\cite{hendrycks2021natural} and VTAB~\cite{zhai2019large}. These datasets are all standard CIL benchmarks and exhibit a significant domain gap compared to ImageNet (used for pretraining). We divided each dataset into T tasks, utilizing the notation `B-$m$ Inc-$n$' to represent class splits, where $m$ denotes the number of classes in the initial task and $n$ indicates the number of classes added per incremental task thereafter. Additionally, to verify NoFA-BC's resistance to forgetting with a large number of incremental tasks, we establish two distinct data splitting setting for ImageNet-R: `B$0$ Inc$20$' and `B$0$ Inc$5$'. We report the mean results and standard deviations of experiments with 5 random seeds. based on results from all seeds. More details are reported in the supplementary material B. \looseness=-1

\noindent\textbf{Evaluation Metrics.} Following the standard protocol established by~\cite{rebuffi2017icarl}, we use two evaluation metrics: the last-task accuracy $A_T$ measuring the performance upon completing all CIL tasks and the average incremental accuracy $\bar{A} = \frac{1}{T} \sum^{T}_{i} A_i$ across all tasks.

\noindent\textbf{Implementation Details.} We use the \textbf{ViT-B/16-IN21K} which is initially pre-trained on ImageNet-21K for all the experiments. For NoFA-BC, we follow the setting of F-OAL~\cite{FOAL2024NeurIPS} to set the expansion dimension $d_B$ of the buffer layer $f_{B}(\cdot)$ to 5000. Other details of hyperparameters of NoFA-BC can be found in Supplementary Material B. 

\subsection{Comparative Study with SOTA Methods}
We compare our method with numerous baselines and state-of-the-art methods and report the results on Table \ref{tab:benchmark} and Figure \ref{fig:benchmark}. We choose Finetune (fine-tuning the entire pretrained model on each task) and Adapter-FT~\cite{chenadaptformer} (fine-tuning only the adapter), L2P~\cite{wang2022learning}, DualPrompt~\cite{wang2022dualprompt}, CODA-Prompt~\cite{smith2023coda}, SimpleCIL~\cite{zhou2024revisiting}, APER~\cite{zhou2024revisiting}, ACMap~\cite{fukuda2024adapter} and EASE~\cite{zhou2024expandable}. Meanwhile, we select ACIL \cite{zhuang2022acil} and DS-AL\cite{zhuang2024ds} as comparative methods among least squares-based approaches. Detailed experimental settings of all methods are reported in supplementary material B.

As shown in Table \ref{tab:benchmark}, our NoFA-BC demonstrates competitive performance across all benchmark datasets for both $\bar{A}$ and $A_T$, with performance gains explicitly marked by red arrows. Notably, NoFA-BC achieves significant improvements in long-sequence scenarios, showing substantial gains on ImageNet-R B0 Inc5. This indicates that NoFA-BC effectively preserve historical knowledge even after numerous tasks. Critically, NoFA-BC demonstrates remarkable superiority over EASE and ACMap on ImageNet-A B0 Inc20, achieving accuracy gains of $9.02\%$ and $7.87\%$ in $A_T$ respectively. This empirically validates the effectiveness of the winner-takes-all allocation strategy and NoFA-BC's robust performance in challenging classification tasks. Also, as shown in Figure \ref{fig:benchmark}, our NoFA-BC leads other compared methods at most time across different datasets and tasks, demonstrating a superior performance of NoFA-BC. 

\begin{table}[t]
    \centering
    \footnotesize
  
    \caption{Ablation study for components on ImageNet-R and ImageNet-A.}

    \label{tab:ablation}
    \begin{tabular}{l|p{1cm}<{\centering}p{1cm}<{\centering}p{1cm}<{\centering}p{1cm}<{\centering}}
        \toprule
        \multirow{2}{2cm}{Modules}& 
                \multicolumn{2}{c}{IN-R B0 Inc5} & \multicolumn{2}{c}{IN-A B0 Inc20} \\
        & 
        $\bar{A}$ & $A_T$ & 
        $\bar{A}$ & $A_T$ \\
        \midrule
            IA          & 78.30 & 70.25 & 68.69 & 57.21 \\
            IA+NFA       & 80.16 & 71.35 & 70.47 & 58.91 \\
            IA+NFA+BLC       & 80.63 & 72.97 & 70.96 & 60.50 \\
            IA+NFA+SE       & 81.09 & 73.42 & 71.55 & 61.71 \\
            IA+NFA+BLC+SE    & \textbf{82.44} & \textbf{74.67} & \textbf{72.46} & \textbf{64.06} \\
        \bottomrule
    \end{tabular}

\end{table}

\subsection{Ablation Study}
\subsubsection{Ablation Study of Proposed Components.} We perform an ablation study of NFA, BLC and SE on ImageNet-R and ImageNet-A. Results are presented in Table \ref{tab:ablation}. Starting from the IA baseline (uniformly fusing features from each adapter), we sequentially incorporate NFA, BLC and SE to validate each component's contribution within the NoFA-BC framework. IA+NFA denotes summing all logits within a specific task from the NFA output as adapters' probability. In Table \ref{tab:ablation}, the results show that incorporating NFA yields a significant improvement over IA alone, and introducing BLC further enhances performance. This demonstrates that the allocation strategy to amplify dominant adapters greatly mitigates catastrophic forgetting. Subsequent utilization of the SE further improves the stability in previous tasks, leading to overall performance gain. These results confirm efficacy of all components within the proposed NoFA-BC.

\begin{table}[t]
    \centering
    
    \begin{minipage}[t]{0.48\textwidth}
        
        \vspace{0pt} 
        \centering
        \caption{Ablation study for different strategies of intra-task competition with NFA.}
        \label{tab:strategy}
        
        {
            \begin{tabular}{l c c c c}
                \toprule
                \multirow{2}{1.7cm}{Intra-task\\Competition} & \multicolumn{2}{c}{IN-R B0 Inc5} & \multicolumn{2}{c}{IN-A B0 Inc20} \\
                \cmidrule(lr){2-3} \cmidrule(lr){4-5}
                & $\bar{A}$ & $A_T$ & $\bar{A}$ & $A_T$ \\
                \midrule
                Weighted\\Sum & 80.16 & 71.35 & 70.47 & 58.91 \\
                Top 50\%     & 80.36 & 71.88 & 70.23 & 58.63 \\
                WTA          & \textbf{80.54} & \textbf{72.43} & \textbf{70.83} & \textbf{59.45} \\
                \bottomrule
            \end{tabular}
        }
    \end{minipage}\hfill 
    \begin{minipage}[t]{0.49\textwidth}
        \vspace{0pt}
        \centering
        \caption{Ablation study of inter-task competition strategy.}
        \label{tab:LOF}
        
        {
            \begin{tabular}{l c c c c}
                \toprule
                \multirow{2}{1.7cm}{Inter-task\\ Competition} & \multicolumn{2}{c}{IN-R B0 Inc5} & \multicolumn{2}{c}{IN-A B0 Inc20} \\
                \cmidrule(lr){2-3} \cmidrule(lr){4-5}
                & $\bar{A}$ & $A_T$ & $\bar{A}$ & $A_T$ \\
                \midrule
                Retain All      & 80.54 & 72.43 & 70.83 & 59.45 \\
                $o=$ 25         & 80.16 & 71.88 & 70.78 & 58.34 \\
                $o=$ 50         & \textbf{80.89} & \textbf{72.97} & \textbf{70.96} & \textbf{60.50} \\
                $o=$ 75         & 80.67 & 72.83 & 70.73 & 60.21 \\
                Highest & 79.32 & 71.03 & 69.88 & 57.63 \\
                \bottomrule
            \end{tabular}
        }
    \end{minipage}
    
\end{table}

\subsubsection{Study on Strategy for Intra-Task Competition.}
In NoFA-BC, after getting the NFA, we employ the WTA to achieve optimal representation of task contribution via intra-task competition. To validate the effectiveness of WTA, we systematically evaluate several alternative approaches: (1)\textbf{Weighted Sum}: Perform summation over probabilities of all classes belonging to a specific task to fully leverage class information from the least squares solution. (2)\textbf{Top 50\%}: Retains the highest-scoring 50\% of logit indices per task, thereby pruning the lower 50\% of category logits to eliminate potential irrelevant knowledge interference. To rigorously investigate impact of extraction strategies, we exclusively employ IA+NFA as baseline, thereby eliminating the influence of inter-task competition and SE. Results of these strategies are presented in Table \ref{tab:strategy}.
Experimental results validate that the WTA achieves optimal results among these strategies. The reason can be that, WTA effectively captures the most task-sensitive information, while the Weighted Sum introduces excessive redundancy by incorporating all class logits, and Top 50\% still fails to achieve substantial improvement due to the introduction of irrelevant knowledge though half are eliminated.

\subsubsection{Study on Strategy for Inter-Task Competition.}
Following the intra-task competition, we further construct the LOF for inter-task competition to reduce the negative effect of less relevant knowledge. In this section, we compare LOF with retaining all adapters and selecting most relevant one. Also, the values of $o$ in LOF are adjusted to modulate the intensity of the inter-task competition. In this experiment, we employ IA+NFA+BLC with WTA as the intra-task competition strategy, eliminating the influence of SE. The results are tabulated in Table \ref{tab:LOF} and `Highest Adapter' denotes retaining only the highest-weighted adapter. The results demonstrate that LOF with $o=50$ achieves optimal performance. When comparing with other strategies, $o=50$ can effectively preserve correct adapters while eliminating erroneous interference. Both the `Highest Adapter' approach and the $o=75$ setting discard more adapters, potentially resulting in excessive loss of valid knowledge and reduced fault tolerance in NoFA-BC.

\subsubsection{Stability Enhancement's impact on Stability and Plasticity.} To intuitively demonstrate SE's impact, we present task-wise classification accuracy visualizations for both old classes (stability) and new classes (plasticity) on ImageNet-R B0 Inc20 and ImageNet-A B0 Inc20 in Figure \ref{fig:bar_sp}. Results reveal that SE significantly enhances performance on old classes across both datasets. While SE's effect on new classes exhibits fluctuations, it remains positive in most cases and performance degradation is small. This observation underscores SE's vital role in boosting stability and overall effectiveness.

\subsubsection{The impact of hyperparameter $\beta$.} 
\begin{figure}[t]
    \centering
    \begin{minipage}[b]{0.44\textwidth}
        \centering
        
        \includegraphics[width=\textwidth, trim=0 15pt 0 10pt, clip]{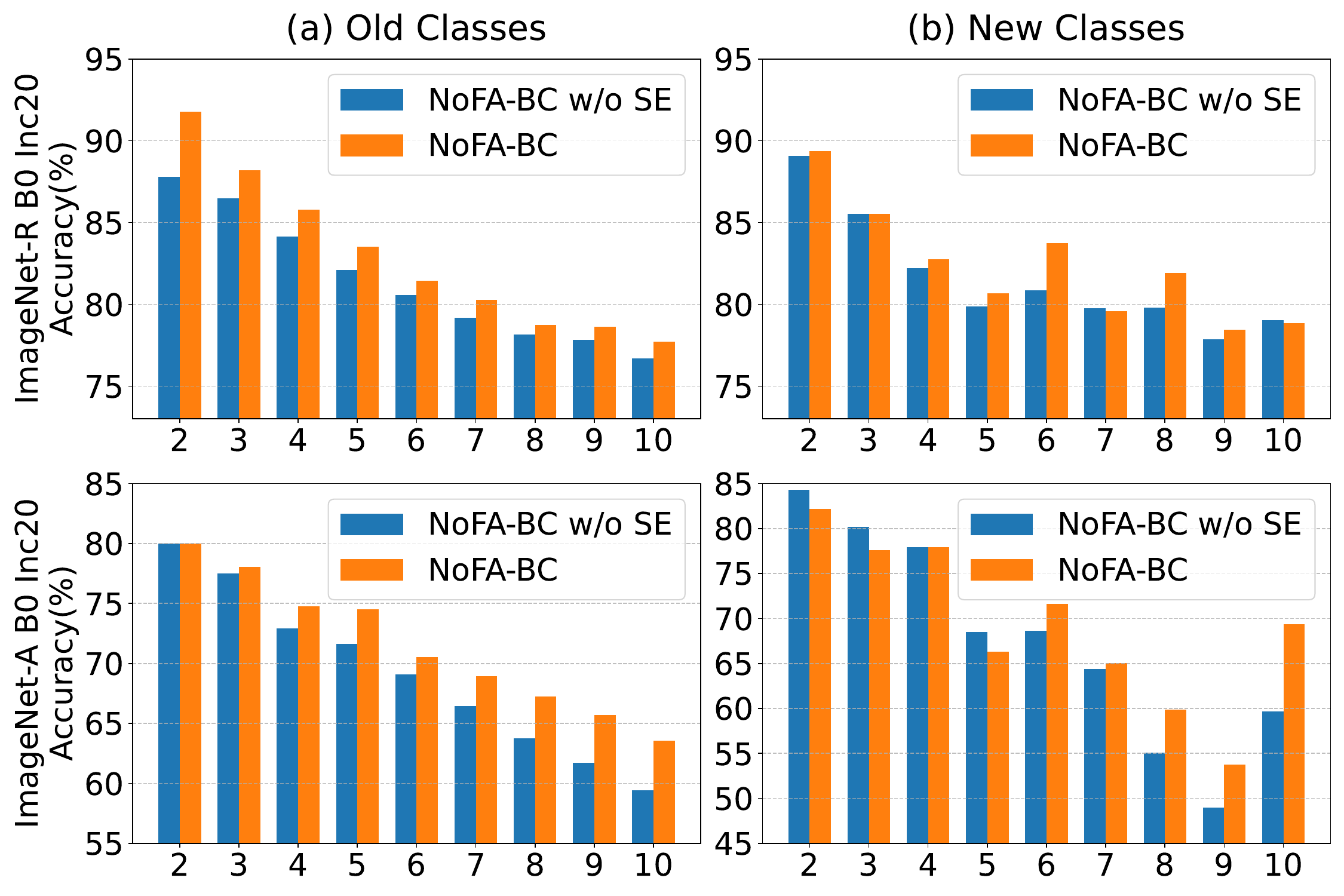}
        \caption{\small Visualization of the impact of Stability Compensation in ImageNet-A B0 Inc20, ImageNet-R B0 Inc20.}
        \label{fig:bar_sp}
    \end{minipage}\hfill
    \begin{minipage}[b]{0.52\textwidth}
        \centering
        \includegraphics[width=\textwidth, trim=0 15pt 0 10pt]{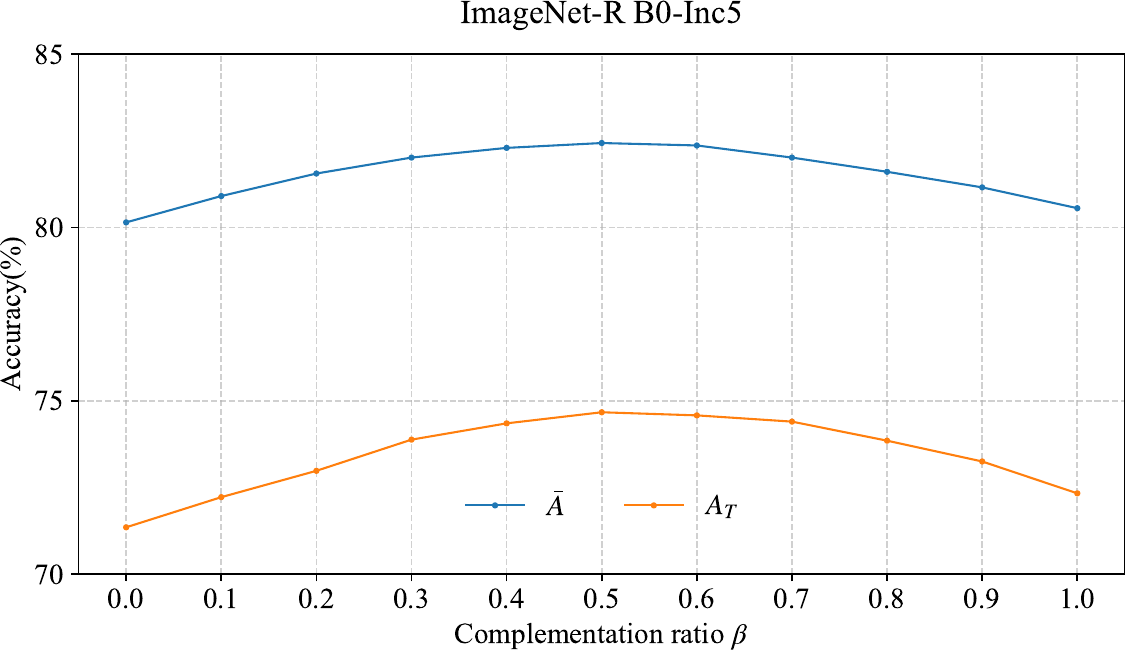}
        \caption{\small The impact of hyperparameter Complementation ratio $\beta$ values in SE on ImageNet-R B0 Inc5.}
        \label{fig:beta}
    \end{minipage}

\end{figure}

$\beta$ serves as a critical hyperparameter in the SE operation, balancing contributions between the NFA and multi-adapters' output. As shown in Figure \ref{fig:beta}, we evaluate $\beta$'s impact on NoFA-BC performance using ImageNet-R B0 Inc5. Results demonstrate performance degradation when $\beta$ approaches 0 or 1. Setting $\beta$ = 0.5 yields optimal performance. 

\subsubsection{Further Analysis on BLC.} 
To further demonstrate the ability of BLC to effectively leverage adapter knowledge while filtering out interference, we designed experiments in Table \ref{table:BLC} including {\bf{A}}: Filter out the correct adapter in BLC with $o=50$; {\bf{B}}: Remove the top 50$\%$ most confident adapters to validate the impact of those excluded by BLC. Experiment {\bf{A}} indicates that adapters within the top 50$\%$ (aside from the correct adapter) also contribute useful features. This suggests that adapters retained by the BLC mechanism have minimal negative impact on the correct adapter while benefiting slightly from useful knowledge provided by other adapters. Experiment {\bf{B}} further confirms our claim: irrelevant adapters introduce negative noise and BLC effectively filters them out.

\subsubsection{Visualization.}
To provide a more intuitive demonstration of the effectiveness of NFA with BLC, we visualize contribution of each adapter when applying NFA with BLC. As shown in Figure \ref{fig:ostrich}, the contribution of each adapter for classifying an ostrich is depicted. NFA with BLC diminishes the weights of adapters unrelated to the task while enhancing those related to it, i.e., adapters containing knowledge about ostriches and related categories (e.g., eagles) are rewarded, while others are suppressed by WTA or even eliminated by LOF. This observation demonstrates that our method can improves the utilization of correct knowledge and effectively mitigates the mutual interference of knowledge. \looseness-1

\begin{table}[t]
    \centering

    \caption{The mechanism of BLC utilizing useful knowledge and suppressing noise.}
    \label{table:BLC}
    
    \footnotesize
    \setlength{\tabcolsep}{5pt}
    \begin{tabular}{ccccc}
        \toprule 
        Settings & {\bf A} & {\bf B} & $o=50$ & correct adapter  \\
        \midrule 
        IN-R Inc5 & 59.14 & 0.12 & 72.97 & 95.93 \\
        IN-A Inc20 & 41.73 & 4.32 & 60.50 & 84.23 \\
        \bottomrule 
    \end{tabular}
    
\end{table}

\subsubsection{Parameter Growth in NoFA-BC.}
\begin{figure}[htb]
    \centering
    \begin{minipage}[t]{0.47\textwidth}
        \vspace{0pt} 
        \centering
        \includegraphics[width=\linewidth]{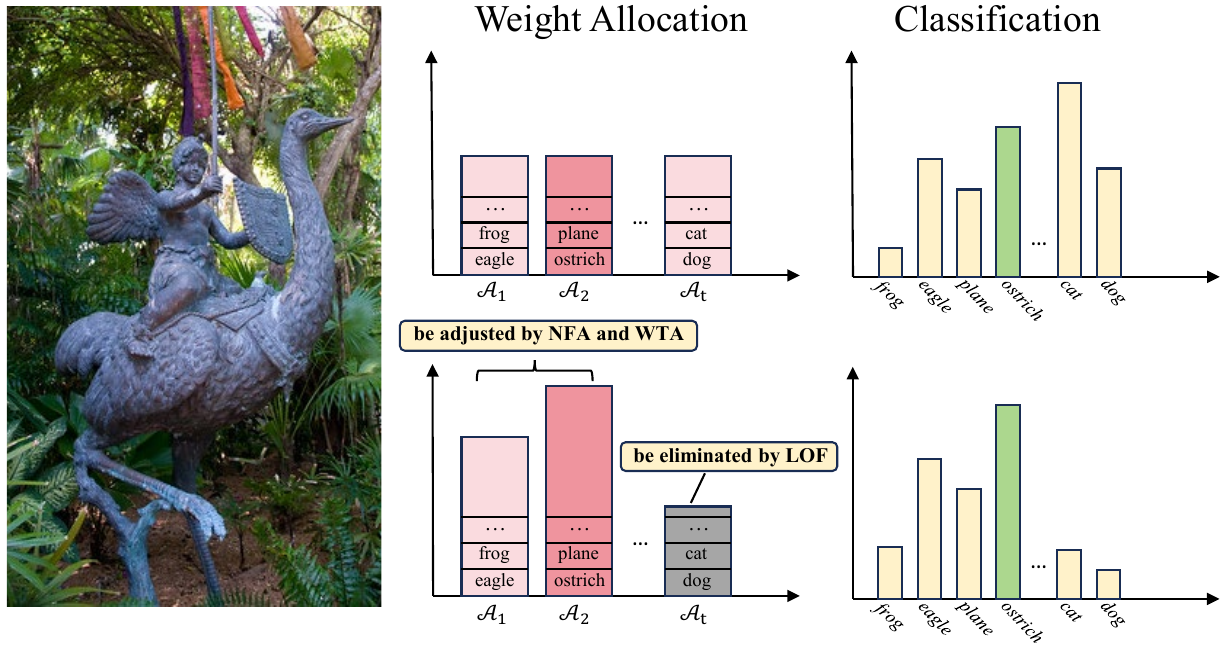} 

        \caption{\small Visualizations of NFA and BLC. Managing the utilization of different adapters improves classification.}
        \label{fig:ostrich}
    \end{minipage}\hfill
    \begin{minipage}[t]{0.52\textwidth}
        
        \vspace{0pt}
        \centering
        \makeatletter\def\@captype{table}\makeatother
        
        \caption{Exploring the impact of the number of adapters on the overall performance of our method in ImageNet-R B0 Inc5.}
        \label{tab:threshold2}
        
        {
            \begin{tabular}{l c c c} 
                \toprule
                {NoFA-BC} & {Params} & $\bar{A}$ & $A_T$ \\
                \midrule
                Only PTM                      & 90.98M  & -     & -     \\
                with 5 adapters  & 95.70M  & 81.59 & 73.87 \\
                with 10 adapters & 101.71M & 82.27 & 74.58 \\
                with 20 adapters & 113.12M & 82.45 & 74.93 \\
                with 40 adapters & 135.28M & 82.44 & 74.67 \\
                \bottomrule
            \end{tabular}
        }
    \end{minipage}
\end{figure}

Our method performs well across diverse scenarios, but adding one adapter per task leads to significant parameter growth. Here, we demonstrates that, NoFA-BC can retain good performance with fewer adapters under ImageNet-R B0-Inc40.  
As shown in Table \ref{tab:threshold2}, we empirically determine an appropriate adapter count to balance capability and parameter growth. Results confirm that using a subset of adapters, combined with NoFA-BC, can still lead to good performance without excessive parameters (e.g., NoFA-BC with 10 adapters v.s. 40 adapters). This observation showcases that our NoFA-BC possesses potential in long-sequence scenarios with fewer storage overhead. 

\section{Conclusion}
\label{sec:conclusion}
{In this paper, we propose the Stability-Compensated Non-Forgetting Allocation (NoFA-BC) for the CIL with PTMs. NoFA-BC integrates a Non-Forgetting Allocator (NFA) with a Bi-Level Competition (BLC) mechanism to deliver accurate and stable knowledge allocation. NFA frames allocation as a recursive least-squares problem, making continual allocator training equivalent to joint learning . Built on NFA, BLC performs intra-task Winner Takes All to capture representative adapter signals and inter-task Last Ones Fall to suppress irrelevant adapters, enabling an effective knowledge allocation management. A Stability Enhancement (SE) process is proposed to further improves the stability. Extensive experiments are conducted to validate NoFA-BC's consistent superiority over existing methods.}

%
%
\bibliographystyle{splncs04}
\bibliography{main}

@String(CVPR= {IEEE Conf. Comput. Vis. Pattern Recog.})

@String(ICCV= {Int. Conf. Comput. Vis.})

@String(ECCV= {Eur. Conf. Comput. Vis.})

@String(IJCAI = {IJCAI})

@String(AAAI = {AAAI})

@String(CVPR  = {CVPR})

@String(ICCV  = {ICCV})

@String(ECCV  = {ECCV})

@inproceedings{zhang2025l3a,
  title     = 	 {{L}3{A}: Label-Augmented Analytic Adaptation for Multi-Label Class Incremental Learning},
  author    =   {Zhang, Xiang and He, Run and Jiao, Chen and Fang, Di and Li, Ming and Zeng, Ziqian and Chen, Cen and Zhuang, Huiping},
  booktitle = 	 {Proceedings of the 42nd International Conference on Machine Learning},
  pages     = 	 {74938--74949},
  year      = 	 {2025},
  editor    = 	 {Singh, Aarti and Fazel, Maryam and Hsu, Daniel and Lacoste-Julien, Simon and Berkenkamp, Felix and Maharaj, Tegan and Wagstaff, Kiri and Zhu, Jerry},
  volume    = 	 {267},
  series    = 	 {Proceedings of Machine Learning Research},
  month     = 	 {13--19 Jul},
  publisher =   {PMLR}
}

@InProceedings{he2025semantic,
  title = 	 {Semantic Shift Estimation via Dual-Projection and Classifier Reconstruction for Exemplar-Free Class-Incremental Learning},
  author =       {He, Run and Fang, Di and Xu, Yicheng and Cui, Yawen and Li, Ming and Chen, Cen and Zeng, Ziqian and Zhuang, Huiping},
  booktitle = 	 {Proceedings of the 42nd International Conference on Machine Learning},
  pages = 	 {22392--22406},
  year = 	 {2025},
  editor = 	 {Singh, Aarti and Fazel, Maryam and Hsu, Daniel and Lacoste-Julien, Simon and Berkenkamp, Felix and Maharaj, Tegan and Wagstaff, Kiri and Zhu, Jerry},
  volume = 	 {267},
  series = 	 {Proceedings of Machine Learning Research},
  month = 	 {13--19 Jul},
  publisher =    {PMLR},
  url = 	 {https://proceedings.mlr.press/v267/he25d.html}
}

@incollection{mccloskey1989catastrophic,
	title={Catastrophic interference in connectionist networks: The sequential learning problem},
	author={McCloskey, Michael and Cohen, Neal J},
	booktitle={Psychology of learning and motivation},
	volume={24},
	pages={109--165},
	year={1989},
	publisher={Elsevier}
}

@article{mermillod2013stability,
  title={The stability-plasticity dilemma: Investigating the continuum from catastrophic forgetting to age-limited learning effects},
  author={Mermillod, Martial and Bugaiska, Aur{\'e}lia and Bonin, Patrick},
  journal={Frontiers in psychology},
  volume={4},
  pages={504},
  year={2013}
}

@inproceedings{rebuffi2017icarl,
  author    = {Rebuffi, Sylvestre-Alvise and Kolesnikov, Alexander and Sperl, Georg and Lampert, Christoph H.},
  title     = {{iCaRL}: Incremental Classifier and Representation Learning},
  booktitle = {Proceedings of the IEEE Conference on Computer Vision and Pattern Recognition (CVPR)},
  month     = jul,
  year      = {2017},
  pages     = {5533--5542},
  doi       = {10.1109/CVPR.2017.587},
  issn      = {1063-6919}
}

@inproceedings{paszke2019pytorch,
	title={Pytorch: An imperative style, high-performance deep learning library},
	author={Paszke, Adam and Gross, Sam and Massa, Francisco and Lerer, Adam and Bradbury, James and Chanan, Gregory and Killeen, Trevor and Lin, Zeming and Gimelshein, Natalia and Antiga, Luca and others},
	booktitle={NeurIPS},
	pages={8026--8037},
	year={2019}
}

@article{belouadah2021comprehensive,
title = {A comprehensive study of class incremental learning algorithms for visual tasks},
journal = {Neural Networks},
volume = {135},
pages = {38-54},
year = {2021},
issn = {0893-6080},
doi = {https://doi.org/10.1016/j.neunet.2020.12.003},
url = {https://www.sciencedirect.com/science/article/pii/S0893608020304202},
author = {Eden Belouadah and Adrian Popescu and Ioannis Kanellos}
}

@inproceedings{liu2020mnemonics,
 author = {Liu, Yaoyao and Su, Yuting and Liu, An-An and Schiele, Bernt and Sun, Qianru},
title = {Mnemonics Training: Multi-Class Incremental Learning Without Forgetting},
booktitle = {Proceedings of the IEEE/CVF Conference on Computer Vision and Pattern Recognition (CVPR)},
month = {June},
year = {2020}
}

@article{masana2022class,
  author={Masana, Marc and Liu, Xialei and Twardowski, Bartłomiej and Menta, Mikel and Bagdanov, Andrew D. and van de Weijer, Joost},
  journal={IEEE Transactions on Pattern Analysis and Machine Intelligence}, 
  title={Class-Incremental Learning: Survey and Performance Evaluation on Image Classification}, 
  year={2023},
  volume={45},
  number={5},
  pages={5513-5533},
  doi={10.1109/TPAMI.2022.3213473}
}

@article{kirkpatrick2017overcoming,
author = {James Kirkpatrick  and Razvan Pascanu  and Neil Rabinowitz  and Joel Veness  and Guillaume Desjardins  and Andrei A. Rusu  and Kieran Milan  and John Quan  and Tiago Ramalho  and Agnieszka Grabska-Barwinska  and Demis Hassabis  and Claudia Clopath  and Dharshan Kumaran  and Raia Hadsell },
title = {Overcoming catastrophic forgetting in neural networks},
journal = {Proceedings of the National Academy of Sciences},
volume = {114},
number = {13},
pages = {3521-3526},
year = {2017},
doi = {10.1073/pnas.1611835114},
URL = {https://www.pnas.org/doi/abs/10.1073/pnas.1611835114}
}

@techreport{krizhevsky2009learning,
	title={Learning multiple layers of features from tiny images},
	author={Krizhevsky, Alex and Hinton, Geoffrey and others},
	year={2009},
}

@inproceedings{yan2021dynamically,
  author    = {Yan, Shipeng and Xie, Jiangwei and He, Xuming},
  title     = {{DER}: Dynamically Expandable Representation for Class Incremental Learning},
  booktitle = {Proceedings of the IEEE/CVF Conference on Computer Vision and Pattern Recognition (CVPR)},
  month     = jun,
  year      = {2021},
  pages     = {3014--3023}
}

@inproceedings{wang2022foster,
  author    = {Wang, Fu-Yun and Zhou, Da-Wei and Ye, Han-Jia and Zhan, De-Chuan},
  no_edito  = {Avidan, Shai and Brostow, Gabriel and Ciss{\'e}, Moustapha and Farinella, Giovanni Maria and Hassner, Tal},
  title     = {{FOSTER}: Feature Boosting and Compression for Class-Incremental Learning},
  booktitle = {Computer Vision -- ECCV 2022},
  year      = {2022},
  publisher = {Springer Nature Switzerland},
  address   = {Cham},
  pages     = {398--414},
  isbn      = {978-3-031-19806-9},
  doi       = {10.1007/978-3-031-19806-9_23}
}

@article{zhao2021memory,
  author={Zhao, Hanbin and Wang, Hui and Fu, Yongjian and Wu, Fei and Li, Xi},
  journal={IEEE Transactions on Neural Networks and Learning Systems}, 
  title={Memory-Efficient Class-Incremental Learning for Image Classification}, 
  year={2022},
  volume={33},
  number={10},
  pages={5966-5977},
  doi={10.1109/TNNLS.2021.3072041}}

@inproceedings{dosovitskiy2020image,
title={An Image is Worth 16x16 Words: Transformers for Image Recognition at Scale},
author={Alexey Dosovitskiy and Lucas Beyer and Alexander Kolesnikov and Dirk Weissenborn and Xiaohua Zhai and Thomas Unterthiner and Mostafa Dehghani and Matthias Minderer and Georg Heigold and Sylvain Gelly and Jakob Uszkoreit and Neil Houlsby},
booktitle={International Conference on Learning Representations},
year={2021},
url={https://openreview.net/forum?id=YicbFdNTTy}
}

@inproceedings{jia2022visual,
author="Jia, Menglin
and Tang, Luming
and Chen, Bor-Chun
and Cardie, Claire
and Belongie, Serge
and Hariharan, Bharath
and Lim, Ser-Nam",
editor="Avidan, Shai
and Brostow, Gabriel
and Ciss{\'e}, Moustapha
and Farinella, Giovanni Maria
and Hassner, Tal",
title="Visual Prompt Tuning",
booktitle="Computer Vision -- ECCV 2022",
year="2022",
publisher="Springer Nature Switzerland",
address="Cham",
pages="709--727",
isbn="978-3-031-19827-4"
}

@inproceedings{rebuffi2017learning,
 author = {Rebuffi, Sylvestre-Alvise and Bilen, Hakan and Vedaldi, Andrea},
 booktitle = {Advances in Neural Information Processing Systems},
 editor = {I. Guyon and U. Von Luxburg and S. Bengio and H. Wallach and R. Fergus and S. Vishwanathan and R. Garnett},
 pages = {},
 publisher = {Curran Associates, Inc.},
 title = {Learning multiple visual domains with residual adapters},
 url = {https://proceedings.neurips.cc/paper_files/paper/2017/file/e7b24b112a44fdd9ee93bdf998c6ca0e-Paper.pdf},
 volume = {30},
 year = {2017}
}

@inproceedings{smith2023coda,
author    = {Smith, James Seale and Karlinsky, Leonid and Gutta, Vyshnavi and Cascante-Bonilla, Paola and Kim, Donghyun and Arbelle, Assaf and Panda, Rameswar and Feris, Rogerio and Kira, Zsolt},
    title     = {{CODA-Prompt}: COntinual Decomposed Attention-Based Prompting for Rehearsal-Free Continual Learning},
    booktitle = {Proceedings of the IEEE/CVF Conference on Computer Vision and Pattern Recognition (CVPR)},
    month     = {June},
    year      = {2023},
    pages     = {11909-11919}
}

@inproceedings{wang2022learning,
    author    = {Wang, Zifeng and Zhang, Zizhao and Lee, Chen-Yu and Zhang, Han and Sun, Ruoxi and Ren, Xiaoqi and Su, Guolong and Perot, Vincent and Dy, Jennifer and Pfister, Tomas},
    title     = {Learning To Prompt for Continual Learning},
    booktitle = {Proceedings of the IEEE/CVF Conference on Computer Vision and Pattern Recognition (CVPR)},
    month     = {June},
    year      = {2022},
    pages     = {139-149}
}

@inproceedings{wang2022dualprompt,
 author="Wang, Zifeng
and Zhang, Zizhao
and Ebrahimi, Sayna
and Sun, Ruoxi
and Zhang, Han
and Lee, Chen-Yu
and Ren, Xiaoqi
and Su, Guolong
and Perot, Vincent
and Dy, Jennifer
and Pfister, Tomas",
editor="Avidan, Shai
and Brostow, Gabriel
and Ciss{\'e}, Moustapha
and Farinella, Giovanni Maria
and Hassner, Tal",
title="DualPrompt: Complementary Prompting for Rehearsal-Free Continual Learning",
booktitle="Computer Vision -- ECCV 2022",
year="2022",
publisher="Springer Nature Switzerland",
address="Cham",
pages="631--648",
isbn="978-3-031-19809-0"
}

@article{thengane2022clip,
  title={CLIP model is an Efficient Continual Learner},
  author={Thengane, Vishal and Khan, Salman and Hayat, Munawar and Khan, Fahad},
  journal={arXiv preprint arXiv:2210.03114},
  year={2022}
}

@article{zhou2024revisiting,
    author = {Zhou, Da-Wei and Cai, Zi-Wen and Ye, Han-Jia and Zhan, De-Chuan and Liu, Ziwei},
    title = {Revisiting Class-Incremental Learning with Pre-Trained Models: Generalizability and Adaptivity are All You Need},
    journal = {International Journal of Computer Vision},
    year = {2024}
}

@inproceedings{chenadaptformer,
 author = {Chen, Shoufa and GE, Chongjian and Tong, Zhan and Wang, Jiangliu and Song, Yibing and Wang, Jue and Luo, Ping},
 booktitle = {Advances in Neural Information Processing Systems},
 editor = {S. Koyejo and S. Mohamed and A. Agarwal and D. Belgrave and K. Cho and A. Oh},
 pages = {16664--16678},
 publisher = {Curran Associates, Inc.},
 title = {AdaptFormer: Adapting Vision Transformers for Scalable Visual Recognition},
 url = {https://proceedings.neurips.cc/paper_files/paper/2022/file/69e2f49ab0837b71b0e0cb7c555990f8-Paper-Conference.pdf},
 volume = {35},
 year = {2022}
}

@inproceedings{hendrycks2021natural,
  author    = {Hendrycks, Dan and Zhao, Kevin and Basart, Steven and Steinhardt, Jacob and Song, Dawn},
    title     = {Natural Adversarial Examples},
    booktitle = {Proceedings of the IEEE/CVF Conference on Computer Vision and Pattern Recognition (CVPR)},
    month     = {June},
    year      = {2021},
    pages     = {15262-15271}
}

@inproceedings{chen2023dynamic,
   author    = {Chen, Xiuwei and Chang, Xiaobin},
    title     = {Dynamic Residual Classifier for Class Incremental Learning},
    booktitle = {Proceedings of the IEEE/CVF International Conference on Computer Vision (ICCV)},
    month     = {October},
    year      = {2023},
    pages     = {18743-18752}
}

@inproceedings{hu2023dense,
    author    = {Hu, Zhiyuan and Li, Yunsheng and Lyu, Jiancheng and Gao, Dashan and Vasconcelos, Nuno},
    title     = {Dense Network Expansion for Class Incremental Learning},
    booktitle = {Proceedings of the IEEE/CVF Conference on Computer Vision and Pattern Recognition (CVPR)},
    month     = {June},
    year      = {2023},
    pages     = {11858-11867}
}

@inproceedings{hendrycks2021many,
  author    = {Hendrycks, Dan and Basart, Steven and Mu, Norman and Kadavath, Saurav and Wang, Frank and Dorundo, Evan and Desai, Rahul and Zhu, Tyler and Parajuli, Samyak and Guo, Mike and Song, Dawn and Steinhardt, Jacob and Gilmer, Justin},
    title     = {The Many Faces of Robustness: A Critical Analysis of Out-of-Distribution Generalization},
    booktitle = {Proceedings of the IEEE/CVF International Conference on Computer Vision (ICCV)},
    month     = {October},
    year      = {2021},
    pages     = {8340-8349}
}

@article{zhai2019large,
  title={A large-scale study of representation learning with the visual task adaptation benchmark},
  author={Zhai, Xiaohua and Puigcerver, Joan and Kolesnikov, Alexander and Ruyssen, Pierre and Riquelme, Carlos and Lucic, Mario and Djolonga, Josip and Pinto, Andre Susano and Neumann, Maxim and Dosovitskiy, Alexey and others},
  journal={arXiv preprint arXiv:1910.04867},
  year={2019}
}

@inproceedings{zhou2024expandable,
     author    = {Zhou, Da-Wei and Sun, Hai-Long and Ye, Han-Jia and Zhan, De-Chuan},
    title     = {Expandable Subspace Ensemble for Pre-Trained Model-Based Class-Incremental Learning},
    booktitle = {Proceedings of the IEEE/CVF Conference on Computer Vision and Pattern Recognition (CVPR)},
    month     = {June},
    year      = {2024},
    pages     = {23554-23564}
}

@article{sun2023pilot,
  title={PILOT: A Pre-Trained Model-Based Continual Learning Toolbox},
  author={Sun, Hai-Long and Zhou, Da-Wei and Ye, Han-Jia and Zhan, De-Chuan},
  journal={arXiv preprint arXiv:2309.07117},
  year={2023}
}

@inproceedings{zhou2024continual,
author = {Zhou, Da-Wei and Sun, Hai-Long and Ning, Jingyi and Ye, Han-Jia and Zhan, De-Chuan},
title = {Continual learning with pre-trained models: a survey},
year = {2024},
isbn = {978-1-956792-04-1},
url = {https://doi.org/10.24963/ijcai.2024/924},
doi = {10.24963/ijcai.2024/924},
booktitle = {Proceedings of the Thirty-Third International Joint Conference on Artificial Intelligence},
articleno = {924},
numpages = {9},
location = {Jeju, Korea},
series = {IJCAI '24}
}

@inproceedings{yue2024mmal,
  author    = {Yue, Xianghu and Zhang, Xueyi and Chen, Yiming and Zhang, Chengwei and Lao, Mingrui and Zhuang, Huiping and Qian, Xinyuan and Li, Haizhou},
  title     = {{MMAL}: Multi-Modal Analytic Learning for Exemplar-Free Audio-Visual Class Incremental Tasks},
  year      = {2024},
  isbn      = {9798400706868},
  publisher = {Association for Computing Machinery},
  address   = {New York, NY, USA},
  no_url    = {https://doi.org/10.1145/3664647.3681607},
  doi       = {10.1145/3664647.3681607},
  booktitle = {Proceedings of the 32nd ACM International Conference on Multimedia},
  pages     = {2428--2437},
  numpages  = {10},
  location  = {Melbourne VIC, Australia},
  series    = {MM '24}
}

@article{zhuang2024ds,
  title   = {{DS-AL}: A Dual-Stream Analytic Learning for Exemplar-Free Class-Incremental Learning},
  volume  = {38},
  no_url  = {https://ojs.aaai.org/index.php/AAAI/article/view/29670},
  doi     = {10.1609/aaai.v38i15.29670},
  number  = {15},
  journal = {Proceedings of the AAAI Conference on Artificial Intelligence},
  author  = {Zhuang, Huiping and He, Run and Tong, Kai and Zeng, Ziqian and Chen, Cen and Lin, Zhiping},
  year    = {2024},
  month   = mar,
  pages   = {17237--17244}
}

@inproceedings{fukuda2024adapter,
   author    = {Fukuda, Takuma and Kera, Hiroshi and Kawamoto, Kazuhiko},
    title     = {Adapter Merging with Centroid Prototype Mapping for Scalable Class-Incremental Learning},
    booktitle = {Proceedings of the IEEE/CVF Conference on Computer Vision and Pattern Recognition (CVPR)},
    month     = {June},
    year      = {2025},
    pages     = {4884-4893}
}

@inproceedings{zhuang2022acil,
  author    = {Zhuang, Huiping and Weng, Zhenyu and Wei, Hongxin and Xie, Renchunzi and Toh, Kar-Ann and Lin, Zhiping},
  title     = {{ACIL}: Analytic Class-Incremental Learning with Absolute Memorization and Privacy Protection},
  booktitle = {Advances in Neural Information Processing Systems},
  no_editor = {S. Koyejo and S. Mohamed and A. Agarwal and D. Belgrave and K. Cho and A. Oh},
  pages     = {11602--11614},
  publisher = {Curran Associates, Inc.},
  volume    = {35},
  year      = {2022},
  no_url    = {https://proceedings.neurips.cc/paper_files/paper/2022/file/4b74a42fc81fc7ee252f6bcb6e26c8be-Paper-Conference.pdf}
}

@inproceedings{xu2024advancing,
  title     = {Advancing Cross-domain Discriminability in Continual Learning of Vision-Language Models},
  author    = {Yicheng Xu and Yuxin Chen and Jiahao Nie and Yusong Wang and Huiping Zhuang and Manabu Okumura},
  booktitle = {The Thirty-eighth Annual Conference on Neural Information Processing Systems},
  year      = {2024},
  no_url    = {https://openreview.net/forum?id=boGxvYWZEq}
}

@article{fang2024air,
  title={Air: Analytic imbalance rectifier for continual learning},
  author={Fang, Di and Zhu, Yinan and Fang, Runze and Chen, Cen and Zeng, Ziqian and Zhuang, Huiping},
  journal={arXiv preprint arXiv:2408.10349},
  year={2024}
}

@article{guo2004pseudoinverse,
title = {A pseudoinverse learning algorithm for feedforward neural networks with stacked generalization applications to software reliability growth data},
journal = {Neurocomputing},
volume = {56},
pages = {101-121},
year = {2004},
issn = {0925-2312},
doi = {https://doi.org/10.1016/S0925-2312(03)00385-0},
url = {https://www.sciencedirect.com/science/article/pii/S0925231203003850},
author = {Ping Guo and Michael R. Lyu}
}

@misc{he2024real,
      title={REAL: Representation Enhanced Analytic Learning for Exemplar-free Class-incremental Learning}, 
      author={Run He and Di Fang and Yizhu Chen and Kai Tong and Cen Chen and Yi Wang and Lap-pui Chau and Huiping Zhuang},
      year={2025},
      eprint={2403.13522},
      archivePrefix={arXiv},
      primaryClass={cs.LG},
      url={https://arxiv.org/abs/2403.13522}, 
}

@inproceedings{gao2025knowledge,
  title={Knowledge memorization and rumination for pre-trained model-based class-incremental learning},
  author={Gao, Zijian and Jia, Wangwang and Zhang, Xingxing and Zhou, Dulan and Xu, Kele and Dawei, Feng and Dou, Yong and Mao, Xinjun and Wang, Huaimin},
  booktitle={Proceedings of the Computer Vision and Pattern Recognition Conference},
  pages={20523--20533},
  year={2025}
}

@inproceedings{FOAL2024NeurIPS,
 author = {Zhuang, Huiping and Liu, Yuchen and He, Run and Tong, Kai and Zeng, Ziqian and Chen, Cen and Wang, Yi and Chau, Lap-Pui},
 booktitle = {Advances in Neural Information Processing Systems},
 editor = {A. Globerson and L. Mackey and D. Belgrave and A. Fan and U. Paquet and J. Tomczak and C. Zhang},
 pages = {41517--41538},
 publisher = {Curran Associates, Inc.},
 title = {F-OAL: Forward-only Online Analytic Learning with Fast Training and Low Memory Footprint in Class Incremental Learning},
 url = {https://proceedings.neurips.cc/paper_files/paper/2024/file/48ffa38c13078d6ce26b328e7f373243-Paper-Conference.pdf},
 volume = {37},
 year = {2024}
}

@InProceedings{LAE2023ICCV,
    author    = {Gao, Qiankun and Zhao, Chen and Sun, Yifan and Xi, Teng and Zhang, Gang and Ghanem, Bernard and Zhang, Jian},
    title     = {A Unified Continual Learning Framework with General Parameter-Efficient Tuning},
    booktitle = {Proceedings of the IEEE/CVF International Conference on Computer Vision (ICCV)},
    month     = {October},
    year      = {2023},
    pages     = {11483-11493}
}

@InProceedings{GKEAL2023CVPR,
    author    = {Zhuang, Huiping and Weng, Zhenyu and He, Run and Lin, Zhiping and Zeng, Ziqian},
    title     = {GKEAL: Gaussian Kernel Embedded Analytic Learning for Few-Shot Class Incremental Task},
    booktitle = {Proceedings of the IEEE/CVF Conference on Computer Vision and Pattern Recognition (CVPR)},
    month     = {June},
    year      = {2023},
    pages     = {7746-7755}
}

@inproceedings{GACL2024NeurIPS,
 author = {Zhuang, Huiping and Chen, Yizhu and Fang, Di and He, Run and Tong, Kai and Wei, Hongxin and Zeng, Ziqian and Chen, Cen},
 booktitle = {Advances in Neural Information Processing Systems},
 doi = {10.52202/079017-2641},
 editor = {A. Globerson and L. Mackey and D. Belgrave and A. Fan and U. Paquet and J. Tomczak and C. Zhang},
 pages = {83024--83047},
 publisher = {Curran Associates, Inc.},
 title = {GACL: Exemplar-Free Generalized Analytic Continual Learning},
 url = {https://proceedings.neurips.cc/paper_files/paper/2024/file/9713d53ee4f31781304b1ca43266f8d1-Paper-Conference.pdf},
 volume = {37},
 year = {2024}
}

@article{AEF-OCLTVT2025,
  author    = {Zhuang, Huiping and Fang, Di and Tong, Kai and Liu, Yuchen and Zeng, Ziqian and Zhou, Xu and Chen, Cen},
  journal   = {IEEE Transactions on Vehicular Technology}, 
  title     = {Online Analytic Exemplar-Free Continual Learning With Large Models for Imbalanced Autonomous Driving Task}, 
  year      = {2025},
  volume    = {74},
  number    = {2},
  pages     = {1949--1958},
  doi       = {10.1109/TVT.2024.3483557},
  ISSN      = {1939-9359},
  month     = feb,
}

@article{HSIC2025JFI,
title = {Class incremental learning with analytic learning for hyperspectral image classification},
journal = {Journal of the Franklin Institute},
volume = {361},
number = {18},
pages = {107285},
year = {2024},
issn = {0016-0032},
doi = {https://doi.org/10.1016/j.jfranklin.2024.107285},
url = {https://www.sciencedirect.com/science/article/pii/S0016003224007063},
author = {Huiping Zhuang and Yue Yan and Run He and Ziqian Zeng}
}

@ARTICLE{CrossACL2025GRSL,
  author={Yan, Yue and Ji, Jianan and Cheng, Yuxuan and Liu, Ye and Xiong, Peiting and Zhuang, Huiping},
  journal={IEEE Geoscience and Remote Sensing Letters}, 
  title={CrossACL: Analytic Continual Learning via Feature Cross for Hyperspectral Image Classification}, 
  year={2025},
  volume={22},
  number={},
  pages={1-5},
  doi={10.1109/LGRS.2025.3587593}}

@article{Cover1965TEC,
  author   = {Cover, Thomas M.},
  journal  = {IEEE Transactions on Electronic Computers},
  title    = {Geometrical and Statistical Properties of Systems of Linear Inequalities with Applications in Pattern Recognition},
  year     = {1965},
  volume   = {EC-14},
  number   = {3},
  pages    = {326--334},
  doi      = {10.1109/PGEC.1965.264137}
}

@article{LwF2017TPAMI,
  title   = {Learning without Forgetting},
  author  = {Li, Zhizhong and Hoiem, Derek},
  journal = {IEEE Transactions on Pattern Analysis and Machine Intelligence},
  year    = {2017},
  volume  = {40},
  number  = {12},
  pages   = {2935--2947},
  doi     = {10.1109/TPAMI.2017.2773081},
  issn    = {1939-3539},
  month   = {dec}
}

\clearpage

\setcounter{page}{1}

\appendix
\setcounter{table}{0}
\setcounter{section}{0}
\setcounter{figure}{0}
\setcounter{equation}{0}

\renewcommand{\thetable}{\Alph{table}}
\renewcommand{\thefigure}{\Alph{figure}}
\renewcommand{\theequation}{\Alph{equation}}
\renewcommand{\thesection}{\Alph{section}}

{\fontfamily{cmr}

\begin{center}
    \large \textbf{Non-Forgetting Knowledge Allocation with Bi-level Competition for
Class-Incremental Learning} \\
\vspace{0.5em}
    \normalsize \textbf{Supplementary Materials}
\end{center}
\vspace{1em} 

In the supplementary materials, we present additional information on NoFA-BC, encompassing more details on implementation and extended experimental results. The supplementary material is organized as follows:
\begin{itemize}

    \item Section \ref{app:Proof} proposes the pseudo-code of NoFA-BC.
    \item Section \ref{app:settings} presents more details of datasets and implementation of each compared method.
    \item Section \ref{app:Rank} demonstrates the impact of projection dimension of the adapters on NoFA-BC.
    \item Section \ref{app:BTW} presents a comparative study on Backward Transfer between NoFA-BC and other methods.
    \item Section \ref{app:Bi} provides further analysis on the Bi-level Competition mechanism. 
    \item Section \ref{app:failure cases} discusses the benefits and risks of knowledge allocation.
    \item Section \ref{app:MOA} presents a comparison between NoFA-BC and recent Mixture of Adapter methods.
    \item Section \ref{app:other} presents the comparison with EASE, ACIL, ACMap, SEMA and MoTE in extra settings.  
    \item Section \ref{app:ref} lists the article references for the Mixture of Adapter methods mentioned in Section \ref{app:MOA}.
\end{itemize}

\section{Pseudo-code of NoFA-BC}\label{app:Proof}
In this section, we present the entire process of NoFA-BC in the form of pseudo-code, including both the training and testing phases.

\begin{algorithm}[h]
    \small
    \caption{NoFA-BC for CIL}
    \label{alg1}
    \textbf{Training:}\\
    \textbf{Input}: Data stream $\{D_t=(\mathcal{X}_t, \mathcal{Y}_t)\}_{t=1}^{T}$, feature expansion projector $f_B(\cdot)$.\\
    \textbf{Initialization}: $\boldsymbol{A}_0\leftarrow \bf0$, $\boldsymbol{C}_0\leftarrow \bf0$
    \begin{algorithmic}[1] 
    \FOR{$t$ in 1 : $T$} 
    \STATE Get the incremental training set $\mathcal{D}_t$.
    \STATE Train task-specific adapter $\mathcal{A}_t$ with $\mathcal{D}_t$.
    \STATE Extract the prototypes ${\bf P}_t = \{\boldsymbol{p}_c|c \in \mathcal{Y}_t\}$ via Eq. \ref{eq:prototypical classifier}.
    \STATE Expand feature to get $\boldsymbol{X}_t$ via Eq. \ref{eq:buffer layer} with $f_B(\cdot)$.
    \STATE Update ${\boldsymbol{A}}_t, {\boldsymbol{C}}_t$ via Eq.\ref{updateAC}.
    \ENDFOR
    \end{algorithmic}
    
    \textbf{Inference after training on task $t$:}\\
    \textbf{Input}: Input image $x$, a pre-trained ViT $f(\cdot)$, the weight of NFA $\boldsymbol{W}^A$, feature expansion projector $f_B(\cdot)$.
    \begin{algorithmic}[1] 
    \STATE Calculate the task weight coefidient  $\{\alpha_1, \alpha_2, ..., \alpha_t\}$ via Eq. \ref{eq:alpha} with $\boldsymbol{W}^A$ and $f_B(\cdot)$.
    \STATE Adjust the allocation of adapters via Bi-level Competition and Eq. \ref{eq:Integration of Adapters}.
    \STATE Do Stability Enhancement via Eq. \ref{eq:Complementary}.
    \end{algorithmic}
\end{algorithm}

\section{Details of Compared Methods}\label{app:settings}

In this section, we provide comprehensive implementation details for all compared methods. All approaches employ \textbf{ViT-B/16-IN21K} as the backbone to ensure a fair comparison. 

For the dataset setting, CIFAR-100 contains 100 classes, ImageNet-R and ImageNet-A each contain 200 classes, and VTAB consists of 50 classes. For all datasets, class order was randomized via a random seed set. We conduct experiments across five distinct random seeds \{1993, 1994, 1995, 1996, 1997\} and report the mean and standard deviation. 

For the implementation of each compared method, we follow the training settings in EASE, using PyTorch~\cite{paszke2019pytorch} to implement all models on a single NVIDIA RTX 4090. SGD with cosine annealing scheduler is chosen for optimizing the adapters. In our method, the learning rate, batch size and training epochs vary across different datasets and the details are depicted in Table \ref{tab:hyper_detail}. Additionally, we follow the setting of F-OAL~\cite{FOAL2024NeurIPS} to set the expansion dimension $d_B$ of the buffer layer $f_{B}(\cdot)$ to 5000 in NoFA-BC. Other method-specific settings are as follows:
\begin{itemize}
\item \textbf{SimpleCIL~\cite{zhou2024revisiting}}. For SimpleCIL, we follow \cite{zhou2024revisiting}'s setting, using solely the pretrained model for feature extraction with cosine similarity computation against class prototypes.

  \begin{table}[h]
        \footnotesize
        \centering
        \setlength{\tabcolsep}{3pt}
        \caption{Details of the training settings for each dataset, based on the configurations provided in PILOT\cite{sun2023pilot}.}
        \label{tab:hyper_detail}
    
        \begin{tabular}{lcccc}
            \toprule
            dataset & batch size & learning rate & weight decay & epochs\\
            \midrule
            \scriptsize{CIFAR-100} & 48 & $2.5\times 10^{-2}$ & $5.0\times 10^{-4}$ & 20\\
            \scriptsize{ImageNet-R} & 16 & $5.0\times 10^{-2}$ & $5.0\times 10^{-3}$ & 20\\
            \scriptsize{ImageNet-A} & 32 & $5.0\times 10^{-2}$ & $5.0\times 10^{-3}$ & 20\\
            \scriptsize{VTAB} & 16 & $3.0\times 10^{-2}$ & $5.0\times 10^{-3}$ & 45\\
            \bottomrule
        \end{tabular}
    \end{table}

\item \textbf{APER-Adapter~\cite{zhou2024revisiting}}: For APER-Adapter, we follow \cite{zhou2024revisiting}'s setting to train a single adapter only on the initial task, freeze it, and utilize it for subsequent feature extraction. The projection dimension $r$ is set to $16$.  \looseness=-1

\item \textbf{L2P~\cite{wang2022learning}}: For L2P, we set the size of prompt pool to $10$, prompt length=$5$, selecting top-5 prompts per instance.

\item \textbf{DualPrompt~\cite{wang2022dualprompt}}: For DualPrompt, we follow \cite{wang2022dualprompt} to set G-Prompt length=$5$ and E-Prompt length=$20$.  \looseness=-1

\item \textbf{CODA-Prompt~\cite{smith2023coda}}: For CODA-Prompt, considering fairness, we allocate $20$ incremental prompts per task (length=$8$) to maintain parameter parity.  \looseness=-1

\item \textbf{ACIL~\cite{zhuang2022acil}}: For ACIL, we set the size of feature expansion matrix $\boldsymbol{B}$ to 5000.

\item \textbf{DS-AL~\cite{zhuang2024ds}}: For DS-AL, we set the compensation ratio $\mathcal{C}$ to 1.0 for all the datasets, with all other hyperparameters aligning with those of ACIL..

\item \textbf{EASE~\cite{zhou2024expandable}}: For EASE, we follow the setting in \cite{zhou2024expandable} to use projection dimension $r=64$ and trade-off parameter $\alpha=0.1$.

\item \textbf{ACMap~\cite{fukuda2024adapter}}: For ACMap, We follow the optimal setting from the paper: no Early Stopping Threshold, with adapter projection dimension $r=64$.
\end{itemize}



\section{Impact of Projection Dimension of Adapter}\label{app:Rank}

For the projection dimension of the adapter, we followed the experimental settings of EASE \cite{zhou2024expandable} by setting the adapter rank $r=64$. However, a larger rank leads to greater parameter growth. With $r=64$, each task's adapter increases the model size by 1.18M parameters. To further explore the impact of the projection dimension, we conducted an extra experiment using different dimension within the adapters. As shown in Table \ref{tab:rank}, we report the perfromance and parameter counts of NoFA-BC with $r=32$, $r=16$, and $r=8$.

The results in Table \ref{tab:rank} indicate that variations in the dimension $r$ have only marginal impact on the performance of NoFA-BC, i.e., the performance consistently maintains with different setting of $r$. This demonstrates that even with relatively small $r$ (e.g., $r=8$), NoFA-BC can still effectively capture task-specific knowledge at each learning phase. This ovservation confirms that, for NoFA-BC, a smaller number of extra parameters can be achieved to significantly control parameter expansion in long-sequence continual learning scenarios.

\section{Study on Backward Transfer}\label{app:BTW}
To more intuitively demonstrate the anti-forgetting capability of NoFA-BC, we evaluated backward transfer (\textit{BTW}) for methods utilizing multiple adapters (EASE \cite{zhou2024expandable}, ACMap \cite{fukuda2024adapter})

\begin{equation}
    \label{eq:BTW}
    BWT=\frac{1}{T-1}\sum_{t=1}^{T-1}A_{T,t}-A_{t,t},
\end{equation}
where $A_{T,t}$ denotes the accuracy achieved by the model when performing inference on data from task $t$ after learning task $T$. The results of BTW are shown in Table \ref{tab:BTW}. The results indicate that NoFA-BC exhibits strong anti-forgetting performance, surpassing EASE and ACMap across all datasets, demonstrating  NoFA-BC's remarkable competitiveness in addressing the core challenge of continual learning, i.e., the catastrophic forgetting.

\begin{table*}[h]
    \centering
     \footnotesize
 
    \caption{Results of BTW in five different datasets.}

    \label{tab:BTW}
    \resizebox{\linewidth}{!}{%
    \begin{tabular}
    {l|p{2.5cm}<{\centering}p{2.5cm}<{\centering}p{2.5cm}<{\centering}p{2.5cm}<{\centering}p{2.5cm}<{\centering}}
        \toprule
        {\multirow{2}{*}{Methods}}& 
                {CIFAR-100 B0 Inc5} & {IN-R B0 Inc5}& {IN-R B0 Inc20} & {IN-A B0 Inc20} & {VTAB B0 Inc10} \\ 
        & 
        $BWT$ & $BWT$ & $BWT$ & $BWT$ & $BWT$
         \\
        \midrule
            EASE    & {-7.14} & {-11.00} & {-7.58} & {-12.18} & {-1.28}\\
            ACMap       & {-7.13} & {-11.34} & {-12.09} & {-12.22} & {-1.74} \\
            NoFA-BC   & \textbf{-3.33} &  \textbf{-6.61} &  \textbf{-6.54} &  \textbf{-9.10} &  \textbf{-0.52}\\
        \bottomrule
    \end{tabular}}

\end{table*}

\section{Further Study on Bi-level Competition}\label{app:Bi}

 In this section, we delve deeper into the mechanism of Bi-level Competition. Both Intra-Task Competition and Inter-Task Competition aim to increase the utilization rate of the adapter corresponding to each sample. To quantify this effect, we introduce a new metric $P_a$, which records the average contributing proportion of the task-specific adapter relative to all adapters of all the samples. That is, after learning all tasks, the sample-wise average of the weighting coefficients of correct adapter.
\begin{equation}
    \label{eq:p_a}
    P_a = \frac{1}{T}\sum_{t=1}^{T} \frac{1}{N_t}\sum_{i}^{N_t} \boldsymbol{p}_{t,i}[t],
\end{equation}
 where $\boldsymbol{p}_{t,i}$ is a coefficient vector that contains the contributions of each adapter and $P_a$ can indicate how the knowledge of corresponding adapter is utilized. The results are summarized in Table \ref{tab:A_a}:

\begin{table}[h]
    \centering
     \footnotesize
     
    \caption{Correct adapter utilization proportions across different Bi-level Competition strategies.}
    
    \label{tab:A_a}
    \resizebox{0.7\linewidth}{!}{%
    \begin{tabular}{l|p{1cm}<{\centering}p{1cm}<{\centering}p{1cm}<{\centering}p{1cm}<{\centering}}
        \toprule
        \multirow{2}{*}{Bi-level Competition}& 
                \multicolumn{2}{c}{IN-R B0 Inc5} & \multicolumn{2}{c}{IN-A B0 Inc20} \\
        & 
        $P_a$ & $A_T$ & 
        $P_a$ & $A_T$ \\
        
        \midrule
        correct adapter & 100.00 & 95.93 & 100.00 & 84.23 \\
        average adapter & 2.50 & 70.83 & 10.00 & 56.12 \\
        \midrule
        \midrule
        
            Weighted Sum          & 17.83 & 71.35 & 27.63 & 58.91 \\
            Top 50\%       & 20.03 & 71.88 & 29.45 & 58.63 \\
            WTA    & \textbf{23.14} & \textbf{72.43} & \textbf{30.61} & \textbf{59.45} \\
            
        \midrule
        \midrule

            WTA + ``$o=$25"    & {24.73} & {72.48} & {30.76} & {59.74} \\
            WTA + ``$o=$50"       & 33.89 & \textbf{72.97} & 35.78 & \textbf{60.50} \\
            WTA + ``$o=$75"         & 25.45 & 72.83 & 34.40 & 60.21 \\
            Highest Adapter   & \textbf{69.24} & 71.03 & \textbf{51.28} & 57.63 \\
        \bottomrule
    \end{tabular}
    }
       
\end{table}

Here, ``correct adapter'' refers to using only the sample-corresponding adapter during the entire inference process (equivalent to knowing the task identity of the sample in advance), and  ``average adapter'' refers to uniformly utilizing all adapters (similar to EASE), without employing NFA and any Bi-level Competition.

In the Intra-task Competition, the WTA strategy outperforms the other approaches, as reflected in both the accuracy metrics in the main paper and the $P_a$ in Table \ref{tab:A_a}. This indicates that the WTA strategy is more effective at capturing task-specific information.

In the Inter-task Competition, experiments demonstrate that the $o=50$ setting optimizes the selection of the correct adapter, thereby improving the accuracy of NoFA-BC. Surprisingly, while the "Highest Adapter" setting achieves the highest $P_a$ score, its accuracy is comparatively lower, contradicting the trend observed in other LOF settings. This occurs because, although the "Highest Adapter" setting eliminates all irrelevant adapters, it also completely eliminates the fault tolerance of NoFA-BC (which relies on the integration of adapters) when an incorrect adapter is selected. The loss of this error-correction mechanism leads to a significant drop in accuracy $A_T$.

\section{Risk of failure cases caused by knowledge allocation}\label{app:failure cases}

Knowledge allocation theoretically introduces certain risks. When the correct knowledge is effectively utilized, task accuracy improves significantly. However, there is also the risk that knowledge may be incorrectly discarded, potentially misleading originally correct classifications toward erroneous outcomes. To investigate this trade-off, we conducted a comparative analysis between NoFA-BC and the "average adapter" strategy (defined in Table \ref{tab:A_a}). We quantified, for each sample, whether the application of knowledge allocation corrected an initially wrong prediction or led a previously correct prediction astray in ImageNet-R and ImageNet-A. The results are as follows in Table \ref{tab：failure}:

\begin{table}[h]
    \centering
    \caption{Observation of samples which are corrected and which are led astray.}
    \label{tab：failure}
    \footnotesize
    \setlength{\tabcolsep}{5pt}
    \resizebox{0.8\linewidth}{!}{%
    \begin{tabular}{ccc}
        \toprule 
        ``average adapter" $\rightarrow$ BLC & correct to wrong &  wrong to correct   \\
        \midrule 
        IN-R Inc5 & 21 & 381\\
        IN-A Inc20 & 17 & 72\\
        \bottomrule 
    \end{tabular}
    }
\end{table}

The results demonstrate that knowledge allocation corrects a substantial number of misclassified samples. While a portion of data indeed becomes incorrectly classified, likely due to the over-utilization of certain erroneous adapters, the benefits of knowledge allocation still far outweigh its risks. This experiment also further confirms the effectiveness of the NFA module from an empirical perspective.

\section{Comparison with other mixture of adapters methods}\label{app:MOA}

Recently, several new works based on the Mixture of Adapters (MOA) paradigm have been published. These methods also employ multiple adapters, utilizing them through different mechanisms:

\begin{table*}[t]
	\caption{\small The average accuracy $\bar{A}$ and the last-task accuracy $A_T$ of Mixture of Adapters methods .The best performance is shown in \textbf{bold} and the second best performance is shown in \underline{underline}. 
	}\label{tab:MOA}
	\centering
    \setlength{\tabcolsep}{1mm}
    \footnotesize
	{
        \resizebox{\linewidth}{!}{%
		\begin{tabular}{p{2.0cm}ccccccccc cccccccc}
			\toprule
			\multicolumn{1}{l}{\multirow{2}{*}{Method}} & 
			\multicolumn{2}{c}{CIFAR B0 Inc5} & \multicolumn{2}{c}{IN-R B0 Inc5} 
			& \multicolumn{2}{c}{IN-R B0 Inc20}
			& \multicolumn{2}{c}{IN-A B0 Inc20}
			& \multicolumn{2}{c}{VTAB B0 Inc10} \\
			& {$\bar{{A}}$} & ${{A}_T}$  
			& {$\bar{{A}}$} & ${{A}_T}$  
			& {$\bar{{A}}$} & ${{A}_T}$ 
			& {$\bar{{A}}$} & ${{A}_T}$ 
			& {$\bar{{A}}$} & ${{A}_T}$ 
			\\
			\midrule
			
		      SEMA 
            &90.27
            &85.23
            &71.22
            &63.25
            &81.64
            &74.52
            &62.97
            &53.32
            &91.23
            &89.68 \\
			MoTE
            & 93.78
            & 88.98
            & 73.21
            & 65.43
            & 82.13
            & 76.39
            & 64.91
            & 57.93
            & 92.77
            & 91.26\\
            \midrule
                \rowcolor{gray!10}
			\textbf{NoFA-BC}
            & \raggedright \bf{94.04
            }
            & \bf{89.21}
            & \bf82.44
            & \bf74.67 
            & \bf83.61
            & \bf77.75   
            & \bf72.51 
            & \bf64.06  
            & \bf{93.98}
            & \bf{93.91} \\
			\bottomrule
		\end{tabular}}
	}
\end{table*}

\begin{figure*}[t]
	\centering
		\includegraphics[width=0.9\columnwidth]{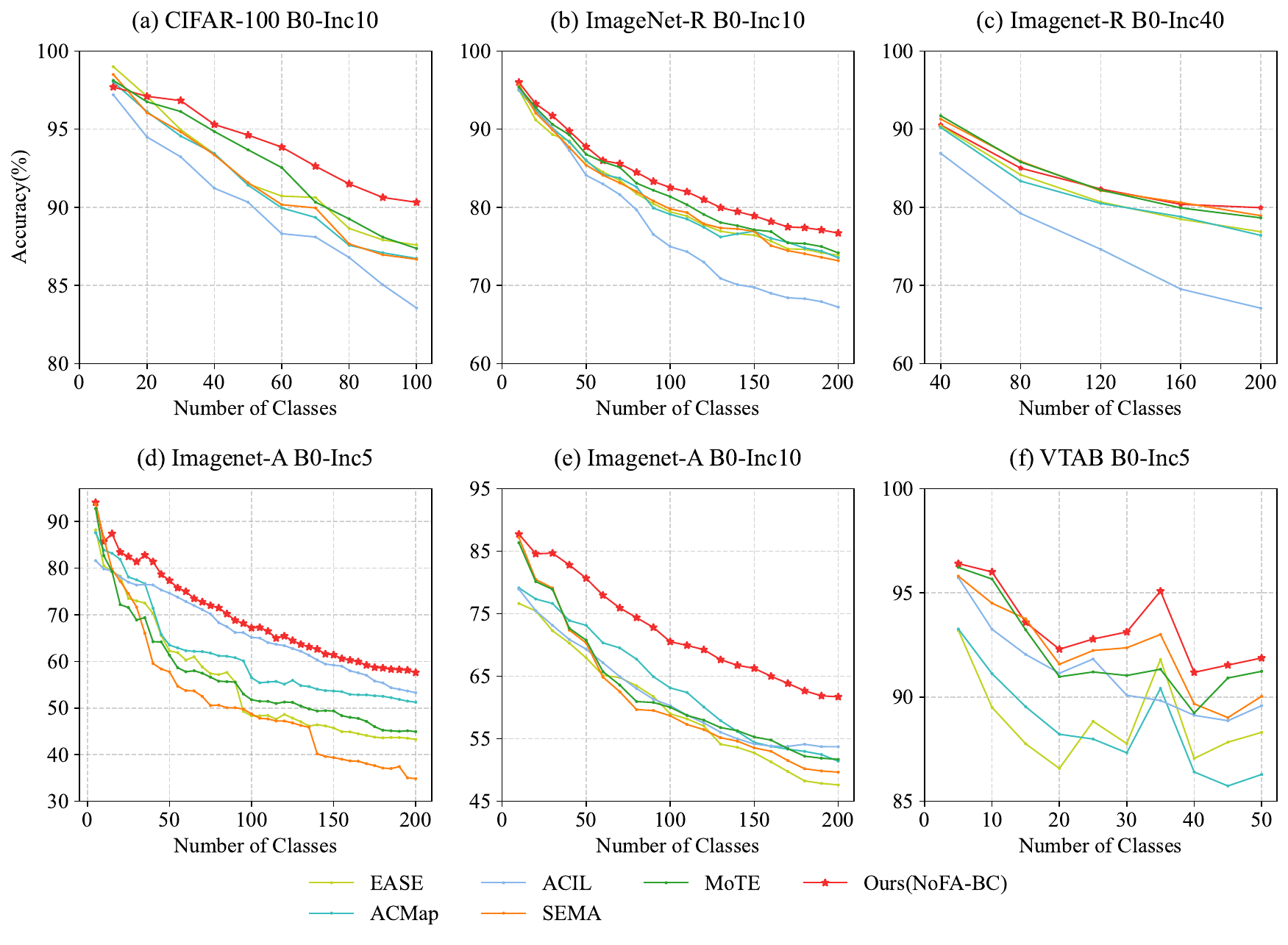}
        
	\caption{\small Accuracy curves of compared methods in dataset settings different from main paper. }
	\label{fig:other}
    
\end{figure*}

\begin{itemize}

    \item \textbf{SEMA} utilizes a self-expansion strategy where representation descriptors detect distribution shifts to automatically trigger the addition of new adapters at specific layers only when necessary.
    \item \textbf{MoTE} employs a training-free inference strategy that filters unreliable adapters based on task scope and fuses the remaining ones using confidence and self-confidence scores.

\end{itemize}

We compared our method against these approaches using the same datasets as in the main results while controlling all hyperparameters (e.g., adapter rank) to ensure a fair comparison. The result is shown in Table \ref{tab:MOA}. Compared to other multi-adapter-based methods, NoFA-BC clearly achieves superior performance in accuracy. Notably, on the two challenging dataset settings: ImageNet-R B0 Inc5 and ImageNet-A B0 Inc20, both SEMA and MoTE exhibit varying degrees of performance degradation, whereas NoFA-BC maintains its superior performance, demonstrating that NoFA-BC utilizes the knowledge in adapters more reasonably.

\section{Extra Comparison of Evaluation Curves}\label{app:other}

In this section, we present accuracy curves of NoFA-BC, ACMap, EASE, ACIL and the MOA methods mentioned before under additional experimental settings. The results in Fig \ref{fig:other} demonstrate that NoFA-BC consistently achieves superior performance.

}

\section{Reference of MOA methods}\label{app:ref}

\begin{itemize}

    \item \textbf{SEMA:} Huiyi Wang, Haodong Lu, Lina Yao, and Dong Gong. Self-expansion of pre-trained models with mixture of adapters for continual learning. In Proceedings of the Computer Vision and Pattern Recognition Conference, pages 10087–10098, 2025.
    \item \textbf{MoTE:} Linjie Li, Zhenyu Wu, and Yang Ji. Mote: Mixture of task-specific experts for pre-trained model-based class-incremental learning. Knowledge-Based Systems, 324:113795, 2025.

\end{itemize}

\end{document}